\documentclass[10pt,twocolumn,letterpaper]{article}

\usepackage[pagenumbers]{wacv}

\usepackage{graphicx}
\usepackage{amsmath}
\usepackage{amssymb}
\usepackage{booktabs}

\usepackage{color}
\usepackage{colortbl}
\usepackage{graphicx}
\usepackage{makecell}
\usepackage{multicol}
\usepackage{multirow}
\usepackage{tabularx}
\usepackage{tikz}

\definecolor{calib3d_purple}{RGB}{112, 48, 160}
\definecolor{calib3d_blue}{RGB}{66, 133, 244}
\definecolor{calib3d_red}{RGB}{231, 66, 52}
\definecolor{calib3d_yellow}{RGB}{251, 189, 5}
\definecolor{calib3d_green}{RGB}{51, 168, 82}
\definecolor{calib3d_gray}{RGB}{165, 165, 165}

\definecolor{cvprblue}{rgb}{0.21,0.49,0.74}

\usepackage[accsupp]{axessibility}
\usepackage{marvosym}

\newcommand\blfootnote[1]{%
\begingroup
\renewcommand\thefootnote{}{}\footnote{#1}%
\addtocounter{footnote}{-1}%
\endgroup
}

\usepackage[pagebackref,breaklinks,colorlinks,citecolor=cvprblue,linkcolor=cvprblue]{hyperref}

\usepackage[capitalize]{cleveref}
\crefname{section}{Sec.}{Secs.}
\Crefname{section}{Section}{Sections}
\Crefname{table}{Table}{Tables}
\crefname{table}{Tab.}{Tabs.}

\usepackage{titletoc}

\usepackage{listings}
\usepackage{color}

\definecolor{codegreen}{rgb}{0,0.6,0}
\definecolor{codegray}{rgb}{0.5,0.5,0.5}
\definecolor{codepurple}{rgb}{0.58,0,0.82}
\definecolor{backcolour}{rgb}{0.95,0.95,0.92}

\def\confName{WACV}
\def\confYear{2025}

\begin{document}
\title{Calib3D: Calibrating Model Preferences for Reliable 3D Scene Understanding}

\author{
    Lingdong Kong$^{1,2,*}$, Xiang Xu$^{3,*}$, Jun Cen$^4$, Wenwei Zhang$^1$, Liang Pan$^1$, Kai Chen$^1$, Ziwei Liu$^{5,\textrm{\Letter}}$
    \\[1ex]
    {\small
    $^1$Shanghai AI Laboratory \quad $^2$National University of Singapore \quad $^3$Nanjing University of Aeronautics and Astronautics
    }
    \\
    {\small
    $^4$The Hong Kong University of Science and Technology \quad $^5$S-Lab, Nanyang Technological University
    }
}

\twocolumn[{%
\renewcommand\twocolumn[1][]{#1}%
\maketitle
\begin{center}
    \centering
    \vspace{-14pt}
    \captionsetup{type=figure}
    \includegraphics[width=\textwidth]{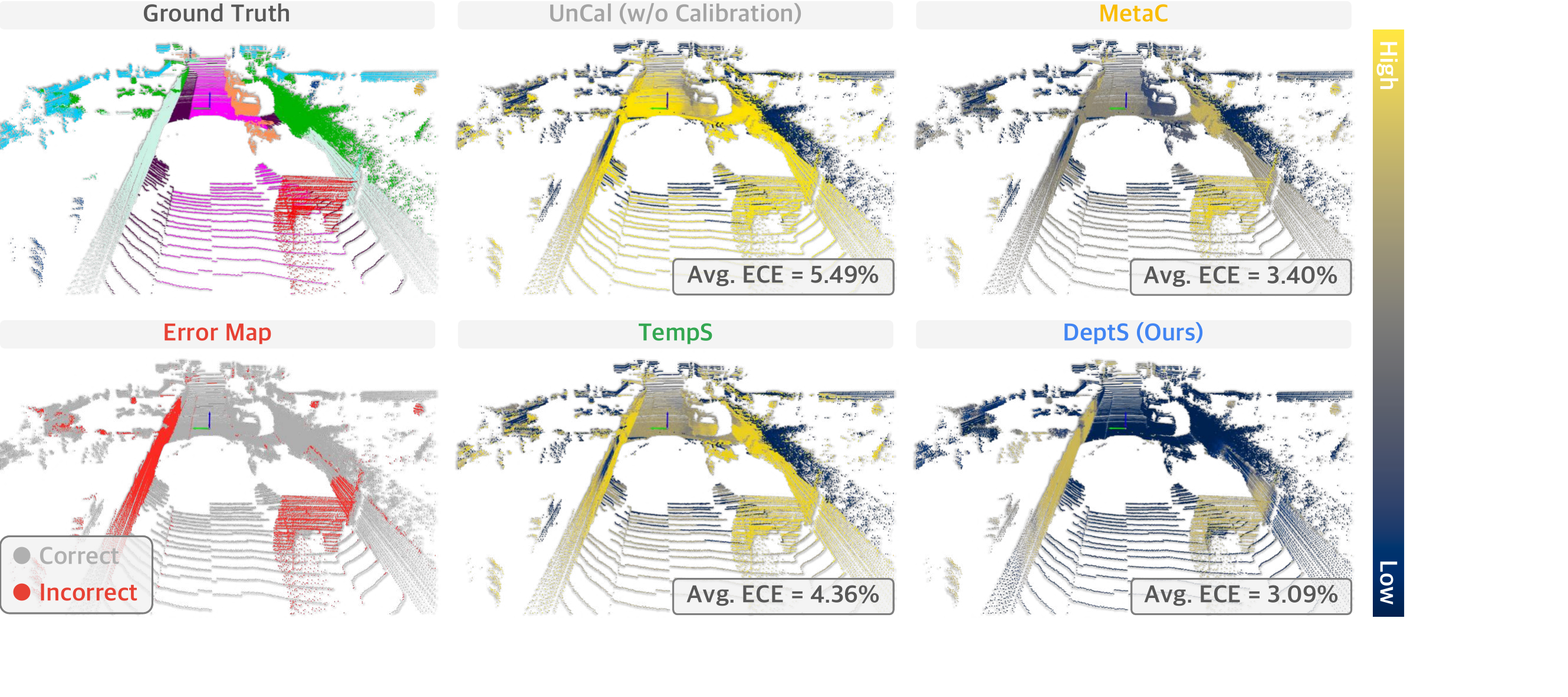}
    \vspace{-16pt}
    \captionof{figure}{
        Well-calibrated 3D scene understanding models are anticipated to deliver \textit{low uncertainties} when predictions are \textit{accurate} and \textit{high uncertainties} when predictions are \textit{inaccurate}. Existing 3D models \cite{zhu2021cylindrical} (\textcolor{calib3d_gray}{UnCal}) and prior calibration methods \cite{guo2017calib,ma2021metaCal} struggled to provide proper uncertainty estimates. Our proposed depth-aware scaling (\textcolor{calib3d_blue}{DeptS}) is capable of outputting accurate estimates, highlighting its potential for real-world usage. The plots shown are the point-wise expected calibration error (ECE) rates. The colormap goes from \textit{dark} to \textit{light}, denoting \textit{low} and \textit{high} error rates, respectively. Best viewed in colors.
    }
    \label{fig:teaser}
\end{center}
}]

%%%%%%%%% FOOTNOTE
\blfootnote{$^{*}$~Lingdong and Xiang contributed equally to this work.}

%%%%%%%%% ABSTRACT
\begin{abstract}
\vspace{-0.1cm}
   Safety-critical 3D scene understanding tasks necessitate not only accurate but also confident predictions from 3D perception models. This study introduces \textbf{Calib3D}, a pioneering effort to benchmark and scrutinize the reliability of 3D scene understanding models from an uncertainty estimation viewpoint. We comprehensively evaluate \textbf{28} state-of-the-art models across \textbf{10} diverse 3D datasets, uncovering insightful phenomena that cope with both the aleatoric and epistemic uncertainties in 3D scene understanding. We discover that despite achieving impressive levels of accuracy, existing models frequently fail to provide reliable uncertainty estimates -- a pitfall that critically undermines their applicability in safety-sensitive contexts. Through extensive analysis of key factors such as network capacity, LiDAR representations, rasterization resolutions, and 3D data augmentation techniques, we correlate these aspects directly with the model calibration efficacy. Furthermore, we introduce \textbf{DeptS}, a novel depth-aware scaling approach aimed at enhancing 3D model calibration. Extensive experiments across a wide range of configurations validate the superiority of our method. We hope this work could serve as a cornerstone for fostering reliable 3D scene understanding. Code and benchmark toolkit are publicly available\footnote{\url{https://github.com/ldkong1205/Calib3D}}.
\vspace{-0.5cm}
\end{abstract}

%%%%%%%%% BODY TEXT
\section{Introduction}
\label{sec:intro}

The reliability of perception systems in real-world conditions is paramount. Safety-critical applications, such as autonomous driving and robot navigation, often rely on robust and accurate predictions from perception models \cite{zhang2023survey,song2024survey,li2024place3d,kong2024lasermix2}.
While learning-based perception models are widely adopted, they often struggle to provide reliable uncertainty estimates \cite{gawlikowski2023survey} and can exhibit over- or under-confidence \cite{hullermeier2021survey}. This poor calibration fails to meet the demands of real-world applications \cite{abdar2021survey,nixon2019measuring,xie2024benchmarking}, contradicting safety requirements in autonomous systems, where precise, confident predictions are critical for obstacle detection \cite{sirohi2022efficientlps,kong2023robo3D,xie2023robobev}. Similar concerns exist in safety-critical areas, \eg, surveillance \cite{kuppers2020multivariate,munir2022towards,popordanoska2024beyond}, healthcare \cite{monteiro2020stochastic,javanbakhat2024assessing,mehrtash2020medical}, and remote sensing \cite{rubwurm2020model,gawlikowski2022advanced}.

Several studies have attempted to understand the reliability of image recognition models and observed insightful phenomenons \cite{kull2019dirichlet,ma2021metaCal,noh2023rankmixup,park2023acls}. Guo \etal \cite{guo2017calib} presented one of the first benchmarks for network calibration, revealing the fact that modern neural networks are no longer well-calibrated. Subsequent works stemmed from similar motivations and drew similar conclusions for other mainstream image-based perception tasks, including object detection \cite{kuppers2020multivariate,munir2022towards,kato2022det3d,oksuz2023towards,popordanoska2024beyond}, depth estimation \cite{kendall2017bayesian,ilg2018uncertainty,poggi2020monocular,upadhyay2022bayescap}, and image segmentation \cite{ding2021local,elias2021calibrated,wang2023selective,mukhoti2018evaluating,bohdal2023label}.

\textbf{Motivation.}
Despite these efforts, the reliability of 3D scene understanding models in providing uncertainty estimates\footnote{In this work, for the sake of clarity, the terms \textit{uncertainty} and \textit{confidence} are used interchangeably, \ie, \textit{uncertainty} $=1~-$ \textit{confidence}.} remains underexplored. 3D data, such as LiDAR and RGB-D camera inputs, are sparser and less structured than images \cite{caesar2020nuScenes,ando2023rangevit,xiao2023survey}. \textbf{Calib3D} is designed to benchmark and study the reliability of 3D models through uncertainty estimation, focusing on both aleatoric and epistemic uncertainties to address real-world, safety-critical challenges. Specifically, our study emphasizes two key aspects:\\
\noindent$\bullet$~\textbf{Aleatoric Uncertainty in 3D.}
We examine how intrinsic factors, \eg, sensor noises \cite{gao2021survey,kong2023robo3D,hao2024mapbench} and point cloud density variations \cite{triess2021survey,liu2024multi,2023CLIP2Scene,chen2023towards}, contribute to data uncertainty in 3D perception, which cannot be reduced by involving more data or using improved models. In Calib3D, we contribute a comprehensive study of \textbf{10} diverse 3D datasets, spanning different sensors, annotations, and scene settings, including driving, off-road, indoor, dynamic, synthetic/simulation, adverse weather conditions, \etc.
    
\noindent$\bullet$~\textbf{Epistemic Uncertainty in 3D.}
Different from the rather unified network structures in 2D \cite{he2016resnet,dosovitskiy2020vit}, 3D scene understanding models encompass diverse structures due to the complex nature of 3D data processing. Our investigation in Calib3D extends to the model uncertainty associated with the diverse 3D architectures, highlighting the importance of addressing knowledge gaps in model training and data representation. A total of \textbf{28} state-of-the-art models are compared and analyzed, shedding light on the future development of more reliable 3D scene understanding models.

Our analysis reveals that while 3D models often achieve high accuracy, their calibration falls short, a gap critical in safety-critical applications. While these models often achieve promising levels of accuracy, their calibration abilities -- essential for trust in safety-critical applications -- consistently fall short of the mark. As shown in \cref{fig:teaser}, better calibration methods are needed to align model confidence with accuracy. Through a detailed examination of network capacity, LiDAR data representations, rasterization, and 3D data augmentation, we identify key areas for improvement.

To further enhance uncertainty estimation capabilities, we propose a depth-aware scaling method called \textbf{DeptS}. Our method is motivated by the observation that uncalibrated models tend to have low accuracy in the middle-to-far region of the ego-vehicle, while, in the meantime, posing severe over-confident predictions. This problem, which is directly correlated with 3D scene structural information, inevitably leads to high calibration errors. To tackle this challenge, we design a depth-correlated temperature to dynamically adjust the logits distribution based on the depth information, exhibiting strong generalizability in calibrating 3D perception models. As demonstrated in \cref{fig:teaser}, DeptS not only significantly improves calibration over uncalibrated models but also outperforms several existing methods \cite{guo2017calib,kull2019dirichlet,wang2023selective,ma2021metaCal}.
To encapsulate, this work is featured by the following seminar contributions:
\\
\noindent$\blacktriangleright$ To the best of our knowledge, \textbf{Calib3D} is the first benchmark dedicated to examining uncertainty in 3D perception models under real-world conditions.
\\
\noindent$\blacktriangleright$ We systematically study \textbf{28} state-of-the-art 3D perception models across \textbf{10} datasets, establishing a foundation for developing more reliable 3D scene understanding models.
\\
\noindent$\blacktriangleright$ We proposed \textbf{DeptS}, a straightforward yet effective depth-aware scaling method that better calibrates the uncertainty estimates for 3D perception models. 
\\
\noindent$\blacktriangleright$ Extensive experimental evaluations across a wide range of 3D datasets/scenarios demonstrate our advantages, shedding light on a more reliable 3D scene understanding that extends well beyond the current state of the art.

\section{Related Work}
\label{sec:related_work}

\noindent\textbf{3D Scene Understanding.}
Holistic 3D perception underpins various real-world applications \cite{caesar2020nuScenes,behley2021semanticKITTI}. Existing methods can be categorized based on 3D representations \cite{uecker2022analyzing}, including range view \cite{milioto2019rangenet++,zhao2021fidnet,cheng2022cenet,xu2020squeezesegv3,kong2023conDA,kong2023rethinking,xu2023frnet}, bird’s eye view \cite{zhou2020polarNet,chen2021polarStream,zhou2021panoptic}, sparse voxel \cite{choy2019minkowski,zhu2021cylindrical,tang2020searching,hong2021dsnet,hong20224dDSNet,cheng2021af2S3Net}, and raw points \cite{hu2020randla,thomas2019kpconv,zhang2023pids,puy23waffleiron}. Recent work combines these representations \cite{liong2020amvNet,xu2021rpvnet,liu2023uniseg} or fuses point clouds with other modalities (\eg, cameras, radars, IMU) \cite{zhuang2021pmf,jaritz2020xMUDA,liu2023seal,peng2023learning,xu2024visual} to enhance accuracy. While 3D perception has progressed on popular benchmarks, the reliability of these models in estimating uncertainty remains unexplored.

\noindent\textbf{Uncertainty Estimation.}
Quantifying uncertainties is crucial in real-world scenarios \cite{ashukha2019pitfalls,hendrycks2016baseline}, especially for safety-critical applications such as 3D scene understanding \cite{ovadia2019uncertainty}. Methods for uncertainty estimation generally fall into various types, \ie, deterministic networks \cite{malinin2018predictive,sensoy2018evidential}, Bayesian methods \cite{denker1990transforming,tishby1989consistent,buntine1991bayesian,hernandez2016alpha,gal2016dropout}, ensembles \cite{lakshminarayanan2017simple,rahaman2021uncertainty,gustafsson2020evaluating}, and test-time augmentations \cite{lyzhov2020greedy,ayhan2022test}. Recent studies pay attention to post-hoc approaches for calibrating uncertainties, which align with practical usages \cite{gawlikowski2023survey}. This work follows this line of research and extends efforts to 3D scene understanding, hoping to enlighten future works on this crucial topic.

\noindent\textbf{Network Calibration.}
As a prevailing research topic, numerous calibration methods have been proposed across various tasks, including image classification \cite{guo2017calib,zhang2021understanding,kull2019dirichlet,ma2021metaCal,noh2023rankmixup,park2023acls,kugathasan2023multiclass}, semantic segmentation \cite{ding2021local,elias2021calibrated,wang2023selective,mukhoti2018evaluating,bohdal2023label,dreissig2023lidarseg}, object detection \cite{kuppers2020multivariate,munir2022towards,kato2022det3d,oksuz2023towards,popordanoska2024beyond}, depth estimation \cite{kendall2017bayesian,ilg2018uncertainty,poggi2020monocular,upadhyay2022bayescap}, remote sensing \cite{rubwurm2020model,gawlikowski2022advanced}, medical imaging \cite{monteiro2020stochastic,javanbakhat2024assessing,mehrtash2020medical}, \etc. However, the calibration of 3D scene understanding models, which relates closely to real-world applications, is rather overlooked in the literature. Dreissig \etal \cite{dreissig2023on} made an initial study of SalsaNext \cite{cortinhal2020salsanext} on the SemanticKITTI \cite{behley2019semanticKITTI} dataset. To our knowledge, \textbf{Calib3D} is the first study of uncertainty estimation for 3D perception models, covering 28 state-of-the-art models across 10 datasets. We also propose \textbf{DeptS}, a novel depth-aware scaling method that effectively improves uncertainty calibration.

\noindent\textbf{3D Robustness.}
The robustness of perception models has gained increasing attention, particularly in driving applications. Research has examined robustness to point cloud corruptions \cite{kong2023robo3D,hahner2021fog,hahner2022snowfall,ren2022modelnet-c,cao2023pasco,yan2024benchmarking,xiao2023semanticSTF}, depth corruptions \cite{kong2023robodepth,gasperini2023md4all,kong2023robodepth_challenge}, and multi-view images \cite{xie2023robobev,huang2024improving,chen2023mbev,hao2024mapbench}. Other works explore robustness against sensor failures \cite{yu2022benchmarking,ge2023metabev} and adversarial attacks \cite{zhang2023survey,xie2024on}. Unlike prior work, our focus is on 3D robustness from the perspective of uncertainty estimation. \textbf{Calib3D} establishes the first comprehensive benchmark in this area, aiming to guide future research in developing more reliable 3D perception models.
\section{Calib3D}
\label{sec:method}

A typical point cloud data $\mathcal{P}=\{\mathbf{p}_i, q_i | i=1, 2, \dots, N\}$ contains $N$ points captured by the sensor, where $\mathbf{p}_i\in\mathbb{R}^{3}$ represents the Cartesian coordinates $(p^x_i,p^y_i,p^z_i)$ and $q_i\in\mathbb{R}^1$ denotes the sensor reflection value, \eg, the laser intensity. For a learning-based system, the data is accompanied by semantic labels $\mathcal{Y}=\{y_i | i=1, 2, \dots, N\}$ for each point in $\mathcal{P}$, with $y_i$ indicating one of $S$ pre-defined semantic classes. The random variables $\mathcal{P}$ and $\mathcal{Y}$ follow a ground truth joint distribution $\pi(\mathcal{P}, \mathcal{Y}) = \pi(\mathcal{Y}|\mathcal{P})\pi(\mathcal{P})$.

\noindent\textbf{Problem Formulation.}
Let $h(\cdot)$ be a 3D model that takes a point cloud $\mathcal{P}$ as the input and outputs class predictions $\hat{Y}=\{\hat{y}_i | i=1, 2, \dots, N\}$ along with confidence scores $\hat{C}=\{\hat{c}_i | i=1, 2, \dots, N\}$, \ie, $h(\mathcal{P})=(\hat{Y},\hat{C})$. Our goal here is two-fold: (1) we want to measure how well the 3D model delivers the uncertainty estimates in its predictions; and (2) we anticipate obtaining a well-calibrated 3D perception model that aligns high confidence scores with accurate predictions. In theory, a perfect model calibration is defined as $\mathbb{P}(\hat{y}_i = y_i | \hat{c}_i = c) = c$, where $c\in[0,1]$ is the expected confidence value. 

\noindent\textbf{Objective.}
To better cater to the real-world requirement, we resort to the non-probabilistic\footnote{In the deep learning context, the \textit{non-probabilistic} output $\mathbf{z}_i$ is often known as \textit{logits}.} output $\mathcal{Z}=\{\mathbf{z}_i | i=1. 2, \dots, N\}$ from the 3D semantic segmentation model for calibration, without altering the model's accuracy. The predicted probability $\hat{c}_i$ can be derived from $\mathbf{z}_i$ using a Softmax function, \ie, $\hat{c}_i=\sigma(\mathbf{z}_i)$, with $\sigma(\cdot)$ denoting the Softmax operation. The overall objective is to produce a calibrated probability $\hat{v}_i$ for each point in $\mathcal{P}$, based on $\hat{y}_i$, $\hat{c}_i$, and $\mathbf{z}_i$. 

\begin{figure}[t]
    \centering
    \includegraphics[width=\linewidth]{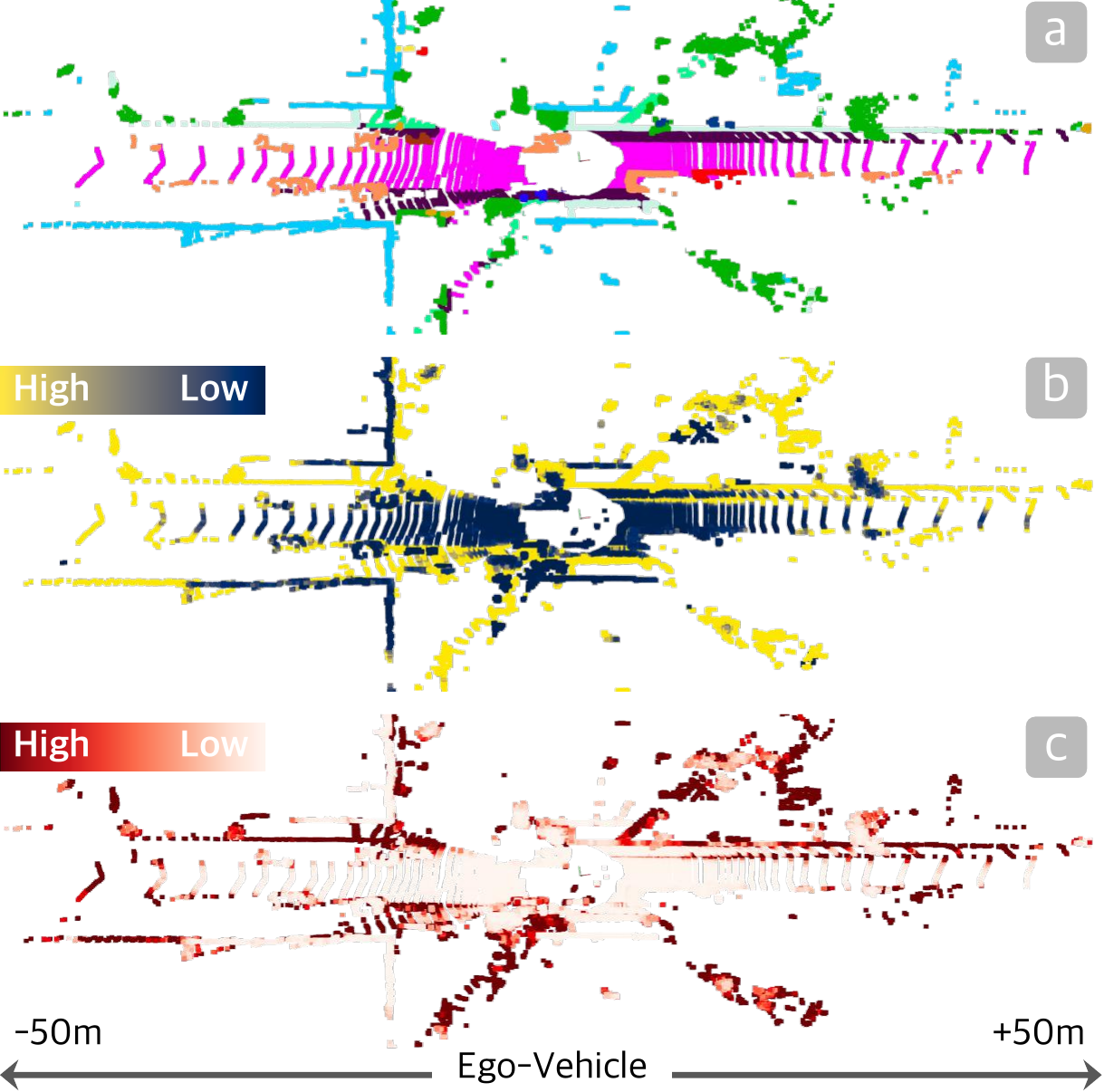}
    \vspace{-0.6cm}
    \caption{Depth-correlated patterns in a $\pm50$m LiDAR-acquired scene from the SemanticKITTI \cite{behley2019semanticKITTI} dataset. (a) Ground truth semantics. (b) Point-wise ECE scores. (c) Point-wise entropy scores.}
    \label{fig:entropy}
    \vspace{-0.3cm}
\end{figure}

\subsection{Calibration Metrics}
\label{sec:calib_metrics}

\noindent\textbf{Expected Calibration Error (ECE).}
Guo \etal \cite{guo2017calib} introduced the ECE metric to assess the confidence calibration of a given neural network. Specifically, ECE measures the difference in expectation between confidence and accuracy:
\vspace{-0.5cm}
\begin{align}
    e_{ece} = \mathbb{E}_{\hat{c}_i}[~ |~ \mathbb{P}(\hat{y}_i = y_i ~|~ \hat{c}_i = c) - c ~| ~]~.
\label{eq:ece}
\end{align}
Based on the definition, a perfectly calibrated model will have an ECE value of zero.

\noindent\textbf{ECE for 3D Scene Understanding.}
In practice, \cref{eq:ece} is approximated by binning continuous variables into equally spaced probability intervals. Different from the conventional image classification task \cite{guo2017calib,ma2021metaCal}, we treat each of the $N$ points\footnote{Note that point clouds may contain varying numbers of points due to acquisitions; we omit such a difference for simplicity.} in $\mathcal{P}$ as unique samples. Assuming a total of $\Tilde{N}$ point clouds in a dataset, we first calculate the ECE score of each point cloud and then average across all point clouds. Such a statistical binning takes the weighted average of the accuracy/confidence difference of each bin as follows:
\begin{align}
    \hat{e}_{ece} = \frac{1}{\Tilde{N}} \sum^{\Tilde{N}}_{\Tilde{n}=1} \sum^M_{m=1} \frac{|B_m|}{N} \left|~ \text{acc}(B_m) - \text{conf}(B_m) ~\right|~,
\label{eq:ece3d}
\end{align}
where $M$ denotes the number of bins used for quantization. $B_m$ denotes the set of samples falling into the $m$-th bin. The difference between $\text{acc}(\cdot)$ and $\text{conf}(\cdot)$ is also known as the \textit{calibration gap} and can be interpreted using reliability diagrams \cite{degroot1983comparison,niculescu2005predicting,guo2017calib}. In the next section, we review the most popular post-hoc methods that have been widely used for calibration in the 2D community.

\subsection{Calibration Methods}

\noindent\textbf{Temperature Scaling (TempS).}
As has been widely verified in theory and practice, a simple extension of the Platt scaling \cite{platt1999probabilistic,niculescu2005predicting} is effective in improving the model calibration. Following \cite{guo2017calib}, a single temperature parameter $T>0$ is used to re-scale the non-probabilistic output $\mathbf{z}_i$:
\begin{align}
    \hat{v}_i^{\text{TempS}} = \max_s \sigma(\frac{\mathbf{z}_i}{T})^{(s)}~,
\label{eq:temps}
\end{align}
where $\sigma(\cdot)$ is the Softmax function, and $\max(\cdot)$ selects the maximum value over $S$ semantic classes. $T$ is learned by minimizing the log-likelihood loss on a validation set.

\noindent\textbf{Logistic Scaling (LogiS).}
A more flexible version of the temperature scaling adopts more complex transformations during re-scaling. Guo \etal \cite{guo2017calib} proposed to use the logistic regression to adjust the non-probabilistic output $\mathbf{z}_i$:
\begin{align}
    \hat{v}_i^{\text{LogiS}} = \max_s \sigma( \mathbf{W}\cdot\mathbf{z}_i + \mathbf{b} )^{(s)}~,
\label{eq:logis}
\end{align}
where $\mathbf{W}$ and $\mathbf{b}$ are optimized based on negative log-likelihood loss on a validation set. In this work, we adopt a vector scaling variant where $\mathbf{W}$ is diagonal.

\noindent\textbf{Dirichlet Scaling (DiriS).}
Assuming the model's outputs follow a Dirichlet distribution (rather than just single probability values), Kull \etal \cite{kull2019dirichlet} further derived the Dirichlet scaling from logistic scaling, which is:
\begin{align}
    \hat{v}_i^{\text{DiriS}} = \max_s \sigma \left( \mathbf{W}\cdot \log (\sigma (\mathbf{z}_i)) + \mathbf{b} \right)^{(s)}~,
\label{eq:diris}
\end{align}
where $\mathbf{W}$ and $\mathbf{b}$ are parameters for a linear parameterization of the predicted probability $\sigma(\mathbf{z}_i)$, and similar to \cref{eq:temps} and \cref{eq:logis}, $\mathbf{W}$ and $\mathbf{b}$ can be optimized based on the negative log-likelihood loss on a validation set.

\noindent\textbf{Meta-Calibration (MetaC).}
Ma \etal \cite{ma2021metaCal} combined a base calibrator (\eg, temperature scaling) with a bipartite ranking model for improved calibration. Specifically, prediction entropy is used to select calibrators; the base calibrator will be used if the entropy is lower than a threshold, and, on the contrary, the predicted output will take random values. Theoretical analyses on high-probability bounds \wrt mis-coverage rate and coverage accuracy are presented in \cite{ma2021metaCal}. In practice, this is formulated as:
\begin{equation}
    \hat{v}_i^{\text{MetaC}} = \begin{cases}
        S^{-1}, & \text{if } -c_i \log(c_i) > \eta \\
        \max_s \sigma(\frac{\mathbf{z}_{i}}{T})^{(s)}, & \text{otherwise} \\
    \end{cases}~,
\label{eq:metac}
\end{equation}
where temperature $T>0$ is learned via log-likelihood minimization on a validation set. $\eta$ is a hand-crafted threshold for filtering high-entropy predictions. It is worth noting that meta-calibration, albeit proven effective in previous literature, will inevitably lose accuracy preservation. The first condition in \cref{eq:metac} introduces randomness to model predictions, which is likely to be impractical regarding real-world, safety-critical applications, \eg, 3D scene understanding.

\subsection{DeptS: Depth-Aware Scaling for 3D Calibration}

\noindent\textbf{Observations.}
While prior calibration methods \cite{platt1999probabilistic,guo2017calib,kull2019dirichlet,ma2021metaCal} have shown appealing calibration performance on image-based perception tasks, their effectiveness on 3D data remains unknown. Unlike RGB images, point cloud data are unordered and texture-less, which inherits extra difficulties in feature learning \cite{gao2021survey,fong2022panoptic-nuScenes,behley2021semanticKITTI}. As shown in \cref{fig:entropy}, we observe a close correlation among calibration error, prediction entropy, and depth -- an inherent 3D information derived from Cartesian coordinates $(p^x_i,p^y_i,p^z_i)$.

\noindent\textbf{Depth Correlations.}
To consolidate this finding, we conduct a quantitative analysis of the relation between calibration error and depth (kindly refer to our Appendix). We calculate the statistics of confidence and accuracy scores of all LiDAR points and then split them into 10 bins based on their depth values, where each bin corresponds to a 5-meter range. We notice from the uncalibrated result that LiDAR points with large depth values (\ie, at the middle-to-far regions of ego-vehicles) tend to have low accuracy. However, the confidence scores of the uncalibrated model do not decrease correspondingly, leading to higher calibration errors. This motivates us to design a method that can resolve the over-confidence issue for LiDAR points with large depths.

\noindent\textbf{Depth-Aware Scaling.}
To fulfill the above pursuit, we propose a simple yet effective depth-aware scaling (DeptS) method for better calibrating 3D scene understanding models. DeptS employs two base calibrators which are selectively used based on prediction entropy calculated using $\hat{c}_i$, which is formulated as follows:
\begin{equation}
    \hat{v}_i^{\text{DeptS}} = \begin{cases}
        \max_s \sigma( \frac{\mathbf{z}_i}{\alpha \cdot T_1} )^{(s)}~, & \text{if } -c_i \log(c_i) > \eta \\
        \max_s \sigma( \frac{\mathbf{z}_i}{\alpha \cdot T_2} )^{(s)}~, & \text{otherwise} \\
    \end{cases}~,
\label{eq:depts}
\end{equation}
where $T_1$ and $T_2$ are temperature parameters and satisfy $T_1>T_2$. The threshold $\eta$ filters high-entropy predictions. Higher entropy indicates a greater likelihood of misclassification, which is often associated with over-confidence \cite{cen2023devil}. Therefore, we use a larger $T_1$ to smooth the logits distribution, which will in turn reduce the confidence score. 

To address the issue of over-confidence for LiDAR points with large depths, we set a depth-correlation coefficient $\alpha$ and use it to re-weight the temperature parameters. The overall process is formulated as follows:
\begin{align}
    \alpha = k_1\cdot d_i + k_2~, ~~~~ d_i = \sqrt{(p^x_i)^2 + (p^y_i)^2 + (p^z_i)^2}~,
\label{eq:depth}
\end{align}
where $k_1$ and $k_2$ are learnable parameters with $k_1>0$. $d_i$ is the depth that is calculated based on the Cartesian coordinates. In this way, LiDAR points with large depth values will have large $\alpha$ values, which are then used to reduce the corresponding confidence score. This in turn mitigates the over-confidence issue for points in the middle-to-far regions. The comparison between our method and the uncalibrated model exhibits the effectiveness of our depth-aware confidence adjustment design. As we will discuss more concretely in the following sections, DeptS contributes a stable improvement in calibrating 3D scene understanding models across a diverse spectrum of 3D datasets.

\begin{table*}[t]
\centering
\caption{The expected calibration error (ECE, the lower the better) of state-of-the-art 3D scene understanding models on the validation sets of the \textit{nuScenes} \cite{fong2022panoptic-nuScenes} and \textit{SemanticKITTI} \cite{behley2019semanticKITTI} datasets. UnCal, TempS, LogiS, DiriS, MetaC, and DeptS denote the uncalibrated, temperature, logistic, Dirichlet, meta, and our proposed depth-aware scaling calibration methods, respectively.}
\vspace{-0.2cm}
\label{tab:benchmark}
\resizebox{0.95\linewidth}{!}{
\begin{tabular}{r|p{1.5cm}<{\raggedleft}|p{1cm}<{\centering}p{1cm}<{\centering}p{1cm}<{\centering}p{1cm}<{\centering}p{1cm}<{\centering}|p{1.3cm}<{\centering}|p{1cm}<{\centering}p{1cm}<{\centering}p{1cm}<{\centering}p{1cm}<{\centering}p{1cm}<{\centering}|p{1.3cm}<{\centering}}
    \toprule
    \multirow{2}{*}{\textbf{Method}} & \multirow{2}{*}{\textbf{Modal}} & \multicolumn{6}{c|}{\textbf{nuScenes} \cite{fong2022panoptic-nuScenes}} & \multicolumn{6}{c}{\textbf{SemanticKITTI} \cite{behley2019semanticKITTI}}
    \\
    & & \textcolor{gray}{UnCal} & TempS & LogiS & DiriS & MetaC & \cellcolor{calib3d_blue!10}\textcolor{calib3d_blue}{DeptS} & \textcolor{gray}{UnCal} & TempS & LogiS & DiriS & MetaC & \cellcolor{calib3d_blue!10}\textcolor{calib3d_blue}{DeptS}
    \\\midrule\midrule
    RangeNet++ \cite{milioto2019rangenet++} & Range~\textcolor{calib3d_blue}{$\bullet$} & \textcolor{gray}{$4.57\%$} & $2.74\%$ & $2.79\%$ & $2.73\%$ & $2.78\%$ & $2.61\%$ & \textcolor{gray}{$4.01\%$} & $3.12\%$ & $3.16\%$ & $3.59\%$ & $2.38\%$ & $2.33\%$
    \\
    SalsaNext \cite{cortinhal2020salsanext} & Range~\textcolor{calib3d_blue}{$\bullet$} & \textcolor{gray}{$3.27\%$} & $2.59\%$ & $2.58\%$ & $2.57\%$ & $2.52\%$ & $2.42\%$ & \textcolor{gray}{$5.37\%$} & $4.29\%$ & $4.31\%$ & $4.11\%$ & $3.35\%$ & $3.19\%$
    \\
    FIDNet \cite{zhao2021fidnet} & Range~\textcolor{calib3d_blue}{$\bullet$} & \textcolor{gray}{$4.89\%$} & $3.35\%$ & $2.89\%$ & $2.61\%$ & $4.55\%$ & $4.33\%$ & \textcolor{gray}{$5.89\%$} & $4.04\%$ & $4.15\%$ & $3.82\%$ & $3.25\%$ & $3.14\%$
    \\
    CENet \cite{cheng2022cenet} & Range~\textcolor{calib3d_blue}{$\bullet$} & \textcolor{gray}{$4.44\%$} & $2.47\%$ & $2.53\%$ & $2.58\%$ & $2.70\%$ & $2.44\%$ & \textcolor{gray}{$5.95\%$} & $3.93\%$ & $3.79\%$ & $4.28\%$ & $3.31\%$ & $3.09\%$
    \\
    RangeViT \cite{ando2023rangevit} & Range~\textcolor{calib3d_blue}{$\bullet$} & \textcolor{gray}{$2.52\%$} & $2.50\%$ & $2.57\%$ & $2.56\%$ & $2.46\%$ & $2.38\%$ & \textcolor{gray}{$5.47\%$} & $3.16\%$ & $4.84\%$ & $8.80\%$ & $3.14\%$ & $3.07\%$
    \\
    RangeFormer \cite{kong2023rethinking} & Range~\textcolor{calib3d_blue}{$\bullet$} & \textcolor{gray}{$2.44\%$} & $2.40\%$ & $2.41\%$ & $2.44\%$ & $2.27\%$ & $2.15\%$ & \textcolor{gray}{$3.99\%$} & $3.67\%$ & $3.70\%$ & $3.69\%$ & $3.55\%$ & $3.30\%$
    \\
    FRNet \cite{xu2023frnet} & Range~\textcolor{calib3d_blue}{$\bullet$} & \textcolor{gray}{$2.27\%$} & $2.24\%$ & $2.22\%$ & $2.28\%$ & $2.22\%$ & $2.17\%$ & \textcolor{gray}{$3.46\%$} & $3.53\%$ & $3.54\%$ & $3.49\%$ & $2.83\%$ & $2.75\%$
    \\\midrule
    PolarNet \cite{zhou2020polarNet} & BEV~\textcolor{calib3d_red}{$\bullet$} & \textcolor{gray}{$4.21\%$} & $2.47\%$ & $2.54\%$ & $2.59\%$ & $2.56\%$ & $2.45\%$ & \textcolor{gray}{$2.78\%$} & $3.54\%$ & $3.71\%$ & $3.70\%$ & $2.67\%$ & $2.59\%$
    \\\midrule
    MinkUNet$_{18}$ \cite{choy2019minkowski} & Voxel~\textcolor{calib3d_yellow}{$\bullet$} & \textcolor{gray}{$2.45\%$} & $2.34\%$ & $2.34\%$ & $2.42\%$ & $2.29\%$ & $2.23\%$ & \textcolor{gray}{$3.04\%$} & $3.01\%$ & $3.08\%$ & $3.30\%$ & $2.69\%$ & $2.63\%$
    \\
    MinkUNet$_{34}$ \cite{choy2019minkowski} & Voxel~\textcolor{calib3d_yellow}{$\bullet$} & \textcolor{gray}{$2.50\%$} & $2.38\%$ & $2.38\%$ & $2.53\%$ & $2.32\%$ & $2.24\%$ & \textcolor{gray}{$4.11\%$} & $3.59\%$ & $3.62\%$ & $3.63\%$ & $2.81\%$ & $2.73\%$
    \\
    Cylinder3D \cite{zhu2021cylindrical} & Voxel~\textcolor{calib3d_yellow}{$\bullet$} & \textcolor{gray}{$3.19\%$} & $2.58\%$ & $2.62\%$ & $2.58\%$ & $2.39\%$ & $2.29\%$ & \textcolor{gray}{$5.49\%$} & $4.36\%$ & $4.48\%$ & $4.42\%$ & $3.40\%$ & $3.09\%$
    \\
    SpUNet$_{18}$ \cite{spconv2022} & Voxel~\textcolor{calib3d_yellow}{$\bullet$} & \textcolor{gray}{$2.58\%$} & $2.41\%$ & $2.46\%$ & $2.59\%$ & $2.36\%$ & $2.25\%$ & \textcolor{gray}{$3.77\%$} & $3.47\%$ & $3.44\%$ & $3.61\%$ & $3.37\%$ & $3.21\%$
    \\
    SpUNet$_{34}$ \cite{spconv2022} & Voxel~\textcolor{calib3d_yellow}{$\bullet$} & \textcolor{gray}{$2.60\%$} & $2.52\%$ & $2.47\%$ & $2.66\%$ & $2.41\%$ & $2.29\%$ & \textcolor{gray}{$4.41\%$} & $4.33\%$ & $4.34\%$ & $4.39\%$ & $4.20\%$ & $4.11\%$
    \\\midrule
    RPVNet \cite{xu2021rpvnet} & Fusion~\textcolor{calib3d_green}{$\bullet$} & \textcolor{gray}{$2.81\%$} & $2.70\%$ & $2.73\%$ & $2.79\%$ & $2.68\%$ & $2.60\%$ & \textcolor{gray}{$4.67\%$} & $4.12\%$ & $4.23\%$ & $4.26\%$ & $4.02\%$ & $3.75\%$
    \\
    2DPASS \cite{yan2022dpass} & Fusion~\textcolor{calib3d_green}{$\bullet$} & \textcolor{gray}{$2.74\%$} & $2.53\%$ & $2.51\%$ & $2.51\%$ & $2.62\%$ & $2.46\%$ & \textcolor{gray}{$2.32\%$} & $2.35\%$ & $2.45\%$ & $2.30\%$ & $2.73\%$ & $2.27\%$
    \\
    SPVCNN$_{18}$ \cite{tang2020searching} & Fusion~\textcolor{calib3d_green}{$\bullet$} & \textcolor{gray}{$2.57\%$} & $2.44\%$ & $2.49\%$ & $2.54\%$ & $2.40\%$ & $2.31\%$ & \textcolor{gray}{$3.46\%$} & $2.90\%$ & $3.07\%$ & $3.41\%$ & $2.36\%$ & $2.32\%$
    \\
    SPVCNN$_{34}$ \cite{tang2020searching} & Fusion~\textcolor{calib3d_green}{$\bullet$} & \textcolor{gray}{$2.61\%$} & $2.49\%$ & $2.54\%$ & $2.61\%$ & $2.37\%$ & $2.28\%$ & \textcolor{gray}{$3.61\%$} & $3.03\%$ & $3.07\%$ & $3.10\%$ & $2.99\%$ & $2.86\%$
    \\
    CPGNet \cite{li2022cpgnet} & Fusion~\textcolor{calib3d_green}{$\bullet$} & \textcolor{gray}{$3.33\%$} & $3.11\%$ & $3.17\%$ & $3.15\%$ & $3.07\%$ & $2.98\%$ & \textcolor{gray}{$3.93\%$} & $3.81\%$ & $3.83\%$ & $3.78\%$ & $3.70\%$ & $3.59\%$
    \\
    GFNet \cite{qiu2022GFNet} & Fusion~\textcolor{calib3d_green}{$\bullet$} & \textcolor{gray}{$2.88\%$} & $2.71\%$ & $2.70\%$ & $2.73\%$ & $2.55\%$ & $2.41\%$ & \textcolor{gray}{$3.07\%$} & $3.01\%$ & $2.99\%$ & $3.05\%$ & $2.88\%$ & $2.73\%$
    \\
    UniSeg \cite{liu2023uniseg} & Fusion~\textcolor{calib3d_green}{$\bullet$} & \textcolor{gray}{$2.76\%$} & $2.61\%$ & $2.63\%$ & $2.65\%$ & $2.45\%$ & $2.37\%$ & \textcolor{gray}{$3.93\%$} & $3.73\%$ & $3.78\%$ & $3.67\%$ & $3.51\%$ & $3.43\%$
    \\\midrule
    KPConv \cite{thomas2019kpconv} & Point~\textcolor{calib3d_gray}{$\bullet$} & \textcolor{gray}{$3.37\%$} & $3.27\%$ & $3.34\%$ & $3.32\%$ & $3.28\%$ & $3.20\%$ & \textcolor{gray}{$4.97\%$} & $4.88\%$ & $4.90\%$ & $4.91\%$ & $4.78\%$ & $4.68\%$
    \\
    PIDS$_{1.25\times}$ \cite{zhang2023pids} & Point~\textcolor{calib3d_gray}{$\bullet$} & \textcolor{gray}{$3.46\%$} & $3.40\%$ & $3.43\%$ & $3.41\%$ & $3.37\%$ & $3.28\%$ & \textcolor{gray}{$4.77\%$} & $4.65\%$ & $4.66\%$ & $4.64\%$ & $4.57\%$ & $4.49\%$
    \\
    PIDS$_{2.0\times}$ \cite{zhang2023pids} & Point~\textcolor{calib3d_gray}{$\bullet$} & \textcolor{gray}{$3.53\%$} & $3.47\%$ & $3.49\%$ & $3.51\%$ & $3.34\%$ & $3.27\%$ & \textcolor{gray}{$4.91\%$} & $4.83\%$ & $4.72\%$ & $4.89\%$ & $4.66\%$ & $4.47\%$ 
    \\
    PTv2 \cite{wu2022ptv2} & Point~\textcolor{calib3d_gray}{$\bullet$} & \textcolor{gray}{$2.42\%$} & $2.34\%$ & $2.46\%$ & $2.55\%$ & $2.48\%$ & $2.19\%$ & \textcolor{gray}{$4.95\%$} & $4.78\%$ & $4.71\%$ & $4.94\%$ & $4.69\%$ & $4.62\%$
    \\
    WaffleIron \cite{puy23waffleiron} & Point~\textcolor{calib3d_gray}{$\bullet$} & \textcolor{gray}{$4.01\%$} & $2.65\%$ & $3.06\%$ & $2.59\%$ & $2.54\%$ & $2.46\%$ & \textcolor{gray}{$3.91\%$} & $2.57\%$ & $2.86\%$ & $2.67\%$ & $2.58\%$ & $2.51\%$
    \\\bottomrule
\end{tabular}}
\vspace{-0.3cm}
\end{table*}

\subsection{Benchmark Configurations}

Our study serves as an early attempt at understanding the predictive preferences of 3D scene understanding models. To consolidate our findings and observations on this topic, we make efforts from the following two aspects in Calib3D.

\noindent\textbf{Aleatoric Uncertainty.}
% 3D data are inherently heterogeneous due to variations in sensor acquisitions, placements, scene conditions, annotation protocols, pre-processing, \etc. A learning-based system trained on diverse data sources tends to exhibit varying confidence and accuracy, particularly under measurement noise. To probe such aleatoric uncertainties, a total of \textbf{10} popular 3D datasets are used in Calib3D, including $^1$\textit{nuScenes} \cite{fong2022panoptic-nuScenes}, $^2$\textit{SemanticKITTI} \cite{behley2019semanticKITTI}, $^3$\textit{Waymo Open} \cite{sun2020waymoOpen}, $^4$\textit{SemanticPOSS} \cite{pan2020semanticPOSS}, $^5$\textit{Synth4D} \cite{saltori2020synth4D}, $^6$\textit{SemanticSTF} \cite{xiao2023semanticSTF}, $^7$\textit{ScribbleKITTI} \cite{unal2022scribbleKITTI}, $^8$\textit{S3DIS} \cite{armeni2016s3dis}, along with the $^9$\textit{nuScenes-C} and $^{10}$\textit{SemanticKITTI-C} datasets from the Robo3D benchmark \cite{kong2023robo3D}. We hope such a comprehensive benchmark study could lay a solid foundation for the future development of reliable 3D scene understanding models. Due to space limits, kindly refer to the Appendix for additional dataset configuration details.
3D data are inherently diverse due to variations in sensor types, placements, and scene conditions. A learning-based system trained on such heterogeneous data often exhibits differing levels of confidence and accuracy, especially under measurement noise. To explore aleatoric uncertainty, Calib3D includes \textbf{10} popular 3D datasets: $^1$\textit{nuScenes} \cite{fong2022panoptic-nuScenes}, $^2$\textit{SemanticKITTI} \cite{behley2019semanticKITTI}, $^3$\textit{Waymo Open} \cite{sun2020waymoOpen}, $^4$\textit{SemanticPOSS} \cite{pan2020semanticPOSS}, $^5$\textit{Synth4D} \cite{saltori2020synth4D}, $^6$\textit{SemanticSTF} \cite{xiao2023semanticSTF}, $^7$\textit{ScribbleKITTI} \cite{unal2022scribbleKITTI}, $^8$\textit{S3DIS} \cite{armeni2016s3dis}, and $^9$\textit{nuScenes-C} and $^{10}$\textit{SemanticKITTI-C} from Robo3D \cite{kong2023robo3D}. This comprehensive study aims to provide a foundation for developing reliable 3D scene understanding models. For additional dataset details, please refer to the Appendix.

\noindent\textbf{Epistemic Uncertainty.}
% The diverse range of 3D scene understanding models introduces multiple factors influencing model uncertainty. To capture these nuances, Calib3D encompasses \textbf{28} state-of-the-art models that have achieved promising accuracy on standard benchmarks. Based on the use of LiDAR representations, we categorize these models into five groups, \ie, $^1$range view \cite{milioto2019rangenet++,cortinhal2020salsanext,zhao2021fidnet,cheng2022cenet,ando2023rangevit,kong2023rethinking,xu2023frnet}, $^2$bird's eye view (BEV) \cite{zhou2020polarNet}, $^3$voxel \cite{choy2019minkowski,zhu2021cylindrical,spconv2022}, $^4$multi-view fusion \cite{xu2021rpvnet,yan2022dpass,tang2020searching,li2022cpgnet,qiu2022GFNet,liu2023uniseg}, and $^5$point-based \cite{thomas2019kpconv,zhang2023pids,wu2022ptv2,puy23waffleiron,qi2017pointnet++,wang2019dgcnn,xu2021paconv} models. Additionally, we investigate the impact of recent 3D data augmentation techniques \cite{xiao2022polarmix,kong2022lasermix,xu2023frnet} and various sparse convolution backends \cite{choy2019minkowski,tang2020searching,spconv2022}, exhibiting design choices that matter for delivering accurate uncertainty estimates. Due to space limits, kindly refer to the Appendix for additional details.
The diverse range of 3D models introduces factors that influence model uncertainty. Calib3D includes \textbf{28} state-of-the-art models with promising performance on standard benchmarks. Based on LiDAR representations, these models are categorized into five groups: $^1$range view \cite{milioto2019rangenet++,cortinhal2020salsanext,zhao2021fidnet,cheng2022cenet,ando2023rangevit,kong2023rethinking,xu2023frnet}, $^2$bird’s eye view (BEV) \cite{zhou2020polarNet}, $^3$voxel \cite{choy2019minkowski,zhu2021cylindrical,spconv2022}, $^4$multi-view fusion \cite{xu2021rpvnet,yan2022dpass,tang2020searching,li2022cpgnet,qiu2022GFNet,liu2023uniseg}, and $^5$point-based models \cite{thomas2019kpconv,zhang2023pids,wu2022ptv2,puy23waffleiron,qi2017pointnet++,wang2019dgcnn,xu2021paconv}. We also examine the impact of 3D data augmentation techniques \cite{xiao2022polarmix,kong2022lasermix,xu2023frnet} and sparse convolution backends \cite{choy2019minkowski,tang2020searching,spconv2022}, identifying key design factors for accurate uncertainty estimates. For additional details, please refer to the Appendix.

\section{Experiments}
\label{sec:experiments}

\subsection{Settings}

\noindent\textbf{Implementation Details.}
The Calib3D benchmark is built using the popular MMDetection3D \cite{mmdet3d} and OpenPCSeg \cite{pcseg2023} codebases, covering a total of 28 models and 10 datasets. We adhere to default configurations for training the models, including the optimizer, learning rate, scheduler, number of training epochs, \etc. Common 3D data augmentations, such as random rotation, flipping, scaling, and jittering, are also applied. For the calibration methods, we follow the conventional setups from prior works. We calculate the predictive entropy statistics for correct/incorrect predictions to select the boundary value as the entropy threshold. Both the proposed DeptS and previous post-hoc calibration methods \cite{guo2017calib,kull2019dirichlet,wang2023selective,ma2021metaCal} are trained under unified configurations. All methods are trained for 20 epochs with a batch size of 8, using the AdamW optimizer \cite{AdamW}. The learning rate is set to $1$e$-3$, and the weight decay is $1$e$-6$. We use four GPUs for both training and evaluation.

\noindent\textbf{Benchmark Protocols.}
To ensure fair comparisons, we unify model training and evaluation configurations during benchmarking. Models are trained on the official \textit{training} split of each dataset and evaluated on the \textit{val} split. We reproduce the originally reported performance without using any extra tricks including test time augmentation, model ensembling, or fine-tuning on validation data. 

\noindent\textbf{Evaluation Metrics.}
The expected calibration error (ECE) metric, as depicted in \cref{eq:ece3d}, is the primary benchmark indicator. We also use class-wise Intersection-over-Union (IoU) and mean IoU (mIoU) to measure 3D segmentation accuracy. For robustness probing, we adopt corruption-wise IoU scores and the mean Resilience Rate (mRR) from Robo3D \cite{kong2023robo3D} to measure the 3D robustness. Kindly refer to the Appendix for more details on these metrics.

\subsection{In-Domain Uncertainty}

\noindent\textbf{Automotive 3D Scenes.}
\cref{tab:benchmark} shows the calibration errors of state-of-the-art 3D scene understanding models on \textit{nuScenes} \cite{fong2022panoptic-nuScenes} and \textit{SemanticKITTI} \cite{behley2019semanticKITTI}. We observe that these models are often poorly calibrated, raising concerns about their reliability in safety-critical contexts. Similar patterns are seen on \textit{Waymo Open} \cite{sun2020waymoOpen} and \textit{SemanticPOSS} \cite{pan2020semanticPOSS} in \cref{tab:in_domain}. Different calibration methods \cite{guo2017calib,kull2019dirichlet,wang2023selective,ma2021metaCal} show promising results in addressing these issues, with temperature scaling \cite{guo2017calib} being particularly effective. Our DeptS sets new benchmarks across all models and datasets, demonstrating the benefits of depth-aware scaling. As highlighted in \cref{fig:teaser}, DeptS holistically improves uncertainty estimation in various regions of LiDAR scenes.

\noindent\textbf{Adverse Weather Conditions.}
The results of PolarNet \cite{zhou2020polarNet}, MinkUNet \cite{choy2019minkowski}, and SPVCNN \cite{tang2020searching} on the \textit{SemanticSTF} \cite{xiao2023semanticSTF} dataset, shown in \cref{tab:in_domain}, underscore the importance of network calibration in 3D scene understanding. Weather conditions pose challenges to accurate uncertainty estimates, complicating real-world applications. Effective calibration with DeptS provides more reliable uncertainty estimations, crucial for safety-critical use cases.

\noindent\textbf{Synthetic LiDAR Data.}
The results on the \textit{Synth4D} \cite{saltori2020synth4D} dataset in \cref{tab:in_domain} suggest that models trained on synthetic data tend to have lower calibration errors, likely due to the less complex nature of simulated point clouds compared to real-world cases. Extra caution is advised when transferring these models to real-world applications.

\noindent\textbf{Sparse Annotations.}
\cref{tab:in_domain} also shows that models trained with weak supervision, such as the line scribbles in \textit{ScribbleKITTI} \cite{unal2022scribbleKITTI}, tend to exhibit higher calibration errors. Compared to dense annotations, weak supervision restricts the model's learning capacity, leading to increased predictive uncertainties. It is thus suggested to adopt calibration methods under such cases to effectively reduce these errors.

\noindent\textbf{Indoor 3D Scenes.}
The last three rows of \cref{tab:in_domain} show calibration errors for PointNet++ \cite{qi2017pointnet++}, DGCNN \cite{wang2019dgcnn}, and PAConv \cite{xu2021paconv} on the \textit{S3DIS} \cite{armeni2016s3dis} dataset. Indoor point clouds also suffer from aleatoric and epistemic uncertainties, emphasizing the importance of network calibration for robust 3D scene understanding. As seen in \cref{tab:benchmark} and \cref{tab:in_domain}, DeptS effectively narrows the gap between confidence and predictive accuracy in these challenging environments.

\begin{figure*}[t]
    \centering
    \includegraphics[width=0.95\linewidth]{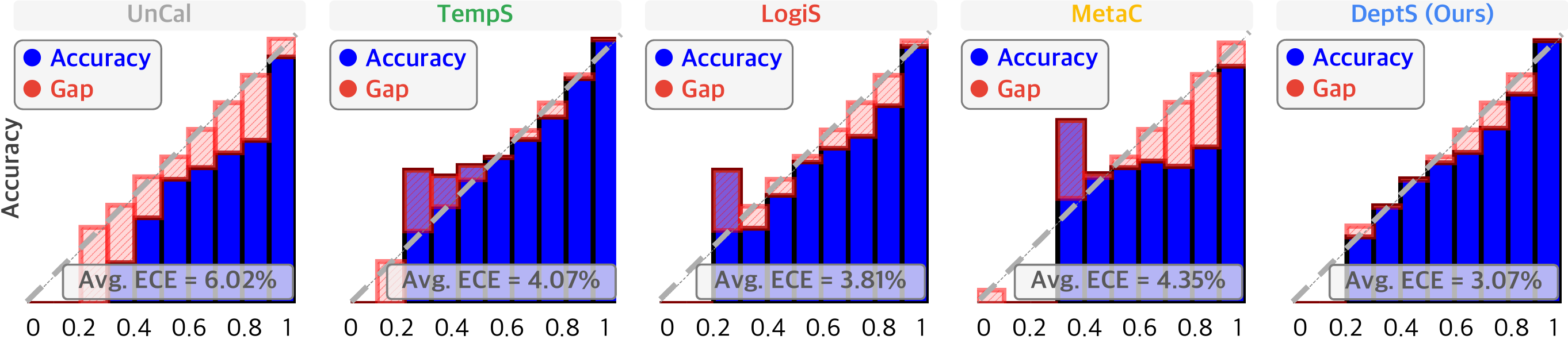}
    \vspace{-0.2cm}
    \caption{The reliability diagrams of visualized calibration gaps from CENet \cite{cheng2022cenet} on \textit{SemanticKITTI} \cite{behley2019semanticKITTI}. UnCal, TempS, LogiS, MetaC, and DeptS denote the uncalibrated, temperature, logistic, meta, and our depth-aware scaling calibration methods, respectively.} 
    \label{fig:diagram}
    \vspace{-0.1cm}
\end{figure*}

\noindent\textbf{Reliability Diagrams.}
As discussed in \cref{sec:calib_metrics}, calibration gaps are well illustrated through reliability diagrams. \cref{fig:diagram} highlights the effectiveness of DeptS in reducing these gaps (depicted in red areas), delivering more accurate uncertainty estimates in practice than prior calibration methods \cite{guo2017calib,ma2021metaCal}. Additional reliability diagrams are in the Appendix.

\begin{table*}[t]
\centering
\caption{The expected calibration error (ECE, the lower the better) and segmentation accuracy (mIoU, the higher the better) of state-of-the-art 3D scene understanding models on the validation sets of \textit{six} heterogeneous benchmarks. UnCal, TempS, LogiS, DiriS, MetaC, and DeptS denote the uncalibrated, temperature, logistic, Dirichlet, meta, and our depth-aware scaling calibration methods, respectively.}
\vspace{-0.2cm}
\label{tab:in_domain}
\scalebox{0.75}{
\begin{tabular}{p{2cm}<{\centering}|p{1.8cm}<{\centering}|p{2.6cm}<{\raggedleft}|p{1.6cm}<{\raggedleft}|p{1.3cm}<{\centering}p{1.3cm}<{\centering}p{1.3cm}<{\centering}p{1.3cm}<{\centering}p{1.3cm}<{\centering}|p{1.6cm}<{\centering}|p{1.6cm}<{\centering}}
    \toprule
    \textbf{Dataset} & \textbf{Type} & \textbf{Method} & \textbf{Modal} & \textcolor{gray}{UnCal} & TempS & LogiS & DiriS & MetaC & \cellcolor{calib3d_blue!10}\textcolor{calib3d_blue}{DeptS} & \textbf{mIoU}
    \\\midrule\midrule
    
    \multirow{3}{*}{\makecell[c]{Waymo Open\\\cite{sun2020waymoOpen}}} & \multirow{3}{*}{High-Res} & PolarNet \cite{zhou2020polarNet} & BEV~\textcolor{calib3d_red}{$\bullet$} & \textcolor{gray}{$3.92\%$} & $1.93\%$ & $1.90\%$ & $1.91\%$ & $2.39\%$ & $1.84\%$ & $58.33\%$
    \\
    & & MinkUNet \cite{choy2019minkowski} & Voxel~\textcolor{calib3d_yellow}{$\bullet$} & \textcolor{gray}{$1.70\%$} & $1.70\%$ & $1.74\%$ & $1.76\%$ & $1.69\%$ & $1.59\%$ & $68.67\%$
    \\
    & & SPVCNN \cite{tang2020searching} & Fusion~\textcolor{calib3d_green}{$\bullet$} & \textcolor{gray}{$1.81\%$} & $1.79\%$ & $1.80\%$ & $1.88\%$ & $1.74\%$ & $1.69\%$ & $68.86\%$
    \\\midrule
    
    \multirow{3}{*}{\makecell[c]{SemanticPOSS\\\cite{pan2020semanticPOSS}}} & \multirow{3}{*}{Dynamic} & PolarNet \cite{zhou2020polarNet} & BEV~\textcolor{calib3d_red}{$\bullet$} & \textcolor{gray}{$4.24\%$} & $8.09\%$ & $7.81\%$ & $8.30\%$ & $5.35\%$ & $4.11\%$ & $52.11\%$
    \\
    & & MinkUNet \cite{choy2019minkowski} & Voxel~\textcolor{calib3d_yellow}{$\bullet$} & \textcolor{gray}{$7.22\%$} & $7.44\%$ & $7.36\%$ & $7.62\%$ & $5.66\%$ & $5.48\%$ & $56.32\%$
    \\
    & & SPVCNN \cite{tang2020searching} & Fusion~\textcolor{calib3d_green}{$\bullet$} & \textcolor{gray}{$8.80\%$} & $6.53\%$ & $6.91\%$ & $7.41\%$ & $4.61\%$ & $3.98\%$ & $53.51\%$
    \\\midrule
    
    \multirow{3}{*}{\makecell[c]{SemanticSTF\\\cite{xiao2023semanticSTF}}} & \multirow{3}{*}{Weather} & PolarNet \cite{zhou2020polarNet} & BEV~\textcolor{calib3d_red}{$\bullet$} & \textcolor{gray}{$5.76\%$} & $4.94\%$ & $4.49\%$ & $4.53\%$ & $4.17\%$ & $4.12\%$ & $51.26\%$
    \\
    & & MinkUNet \cite{choy2019minkowski} & Voxel~\textcolor{calib3d_yellow}{$\bullet$} & \textcolor{gray}{$5.29\%$} & $5.21\%$ & $4.96\%$ & $5.10\%$ & $4.78\%$ & $4.72\%$ & $50.22\%$
    \\
    & & SPVCNN \cite{tang2020searching} & Fusion~\textcolor{calib3d_green}{$\bullet$} & \textcolor{gray}{$5.85\%$} & $5.53\%$ & $5.16\%$ & $5.05\%$ & $5.12\%$ & $4.97\%$ & $51.73\%$
    \\\midrule
    
    \multirow{3}{*}{\makecell[c]{ScribbleKITTI\\\cite{unal2022scribbleKITTI}}} & \multirow{3}{*}{Scribble} & PolarNet \cite{zhou2020polarNet} & BEV~\textcolor{calib3d_red}{$\bullet$} & \textcolor{gray}{$4.65\%$} & $4.59\%$ & $4.56\%$ & $4.55\%$ & $3.25\%$ & $3.09\%$ & $55.22\%$
    \\
    & & MinkUNet \cite{choy2019minkowski} & Voxel~\textcolor{calib3d_yellow}{$\bullet$} & \textcolor{gray}{$7.97\%$} & $7.13\%$ & $7.29\%$ & $7.21\%$ & $5.93\%$ & $5.74\%$ & $59.87\%$
    \\
    & & SPVCNN \cite{tang2020searching} & Fusion~\textcolor{calib3d_green}{$\bullet$} & \textcolor{gray}{$7.04\%$} & $6.63\%$ & $6.93\%$ & $6.66\%$ & $5.34\%$ & $5.13\%$ & $60.22\%$
    \\\midrule
    
    \multirow{3}{*}{\makecell[c]{Synth4D\\\cite{saltori2020synth4D}}} & \multirow{3}{*}{Synthetic} & PolarNet \cite{zhou2020polarNet} & BEV~\textcolor{calib3d_red}{$\bullet$} & \textcolor{gray}{$1.68\%$} & $0.93\%$ & $0.75\%$ & $0.72\%$ & $1.54\%$ & $0.69\%$ & $85.63\%$
    \\
    & & MinkUNet \cite{choy2019minkowski} & Voxel~\textcolor{calib3d_yellow}{$\bullet$} & \textcolor{gray}{$2.43\%$} & $2.72\%$ & $2.43\%$ & $2.05\%$ & $4.01\%$ & $2.39\%$ & $69.11\%$
    \\
    & & SPVCNN \cite{tang2020searching} & Fusion~\textcolor{calib3d_green}{$\bullet$} & \textcolor{gray}{$2.21\%$} & $2.35\%$ & $1.86\%$ & $1.70\%$ & $3.44\%$ & $1.67\%$ & $69.68\%$
    \\\midrule
    
    \multirow{3}{*}{\makecell[c]{S3DIS\\\cite{armeni2016s3dis}}} & \multirow{3}{*}{Indoor} & PointNet++ \cite{qi2017pointnet++} & Point~\textcolor{calib3d_gray}{$\bullet$} & \textcolor{gray}{$9.13\%$}  & $8.36\%$ & $7.83\%$ & $8.20\%$ & $6.93\%$ & $6.79\%$ & $56.96\%$
    \\
    &  & DGCNN \cite{wang2019dgcnn} & Point~\textcolor{calib3d_gray}{$\bullet$} & \textcolor{gray}{$6.00\%$} & $6.23\%$ & $6.35\%$ & $7.12\%$ & $5.47\%$ & $5.39\%$ & $54.50\%$
    \\
    &  & PAConv \cite{xu2021paconv} & Point~\textcolor{calib3d_gray}{$\bullet$} & \textcolor{gray}{$8.38\%$} & $5.87\%$ & $6.03\%$ & $5.98\%$ & $4.67\%$ & $4.57\%$ & $66.60\%$
    \\\bottomrule
\end{tabular}}
\vspace{-0.2cm}
\end{table*}

\begin{table*}[t]
\centering
\caption{The expected calibration error (ECE the lower the better) of the MinkUNet \cite{choy2019minkowski} model under eight domain-shift scenarios from the \textit{nuScenes-C} and \textit{SemanticKITTI-C} datasets in the \textit{Robo3D} benchmark \cite{kong2023robo3D}. UnCal, TempS, LogiS, DiriS, MetaC, and DeptS denote the uncalibrated, temperature, logistic, Dirichlet, meta, and our depth-aware scaling calibration methods, respectively.}
\vspace{-0.2cm}
\label{tab:robo3d}
\scalebox{0.75}{
\begin{tabular}{r|p{1.18cm}<{\centering}<{\centering}<{\raggedleft}p{1.18cm}<{\centering}<{\centering}<{\raggedleft}p{1.18cm}<{\centering}<{\centering}<{\raggedleft}p{1.18cm}<{\centering}<{\centering}<{\raggedleft}p{1.18cm}<{\centering}<{\centering}<{\raggedleft}|p{1.4cm}<{\centering}<{\raggedleft}|p{1.18cm}<{\centering}<{\centering}<{\raggedleft}p{1.18cm}<{\centering}<{\centering}<{\raggedleft}p{1.18cm}<{\centering}<{\centering}<{\raggedleft}p{1.18cm}<{\centering}<{\centering}<{\raggedleft}p{1.18cm}<{\centering}<{\centering}<{\raggedleft}|p{1.4cm}<{\centering}<{\raggedleft}}
\toprule
\multirow{2}{*}{\textbf{Type}} & \multicolumn{6}{c|}{\textbf{nuScenes-C}} & \multicolumn{6}{c}{\textbf{SemanticKITTI-C}}
\\
& \textcolor{gray}{UnCal} & TempS & LogiS & DiriS & MetaC & \cellcolor{calib3d_blue!10}\textcolor{calib3d_blue}{DeptS} & \textcolor{gray}{UnCal} & TempS & LogiS & DiriS & MetaC & \cellcolor{calib3d_blue!10}\textcolor{calib3d_blue}{DeptS}
\\\midrule\midrule
\rowcolor{calib3d_green!8}\textcolor{calib3d_green}{\textbf{Clean}}~\textcolor{calib3d_green}{$\bullet$} & \textcolor{calib3d_green}{$2.45\%$} & \textcolor{calib3d_green}{$2.34\%$} & \textcolor{calib3d_green}{$2.34\%$} & \textcolor{calib3d_green}{$2.42\%$} & \textcolor{calib3d_green}{$2.29\%$} & \textcolor{calib3d_green}{$2.23\%$} & \textcolor{calib3d_green}{$3.04\%$} & \textcolor{calib3d_green}{$3.01\%$} & \textcolor{calib3d_green}{$3.08\%$} & \textcolor{calib3d_green}{$3.30\%$} & \textcolor{calib3d_green}{$2.69\%$} & \textcolor{calib3d_green}{$2.63\%$}
\\\midrule
Fog~\textcolor{calib3d_red}{$\circ$} & \textcolor{gray}{$5.52\%$} & $5.42\%$ & $5.49\%$ & $5.43\%$ & $4.77\%$ & $4.72\%$ & \textcolor{gray}{$12.66\%$} & $12.55\%$ & $12.67\%$ & $12.48\%$ & $11.08\%$ & $10.94\%$
\\
Wet Ground~\textcolor{calib3d_red}{$\circ$} & \textcolor{gray}{$2.63\%$} & $2.54\%$ & $2.54\%$ & $2.64\%$ & $2.55\%$ & $2.52\%$ & \textcolor{gray}{$3.55\%$} & $3.46\%$ & $3.54\%$ & $3.72\%$ & $3.33\%$ & $3.28\%$
\\
Snow~\textcolor{calib3d_red}{$\circ$} & \textcolor{gray}{$13.79\%$} & $13.32\%$ & $13.53\%$ & $13.59\%$ & $11.37\%$ & $11.31\%$ & \textcolor{gray}{$7.10\%$} & $6.96\%$ & $6.95\%$ & $7.26\%$ & $5.99\%$ & $5.63\%$
\\
Motion Blur~\textcolor{calib3d_red}{$\circ$} & \textcolor{gray}{$9.54\%$} & $9.29\%$ & $9.37\%$ & $9.01\%$ & $8.32\%$ & $8.29\%$ & \textcolor{gray}{$11.31\%$} & $11.16\%$ & $11.24\%$ & $12.13\%$ & $9.00\%$ & $8.97\%$
\\
Beam Missing~\textcolor{calib3d_red}{$\circ$} & \textcolor{gray}{$2.58\%$} & $2.48\%$ & $2.49\%$ & $2.57\%$ & $2.53\%$ & $2.47\%$ & \textcolor{gray}{$2.87\%$} & $2.83\%$ & $2.84\%$ & $2.98\%$ & $2.83\%$ & $2.79\%$
\\
Crosstalk~\textcolor{calib3d_red}{$\circ$} & \textcolor{gray}{$13.64\%$} & $13.00\%$ & $12.97\%$ & $13.44\%$ & $9.98\%$ & $9.73\%$ & \textcolor{gray}{$4.93\%$} & $4.83\%$ & $4.86\%$ & $4.81\%$ & $3.54\%$ & $3.48\%$
\\
Incomplete Echo~\textcolor{calib3d_red}{$\circ$} & \textcolor{gray}{$2.44\%$} & $2.33\%$ & $2.33\%$ & $2.42\%$ & $2.32\%$ & $2.21\%$ & \textcolor{gray}{$3.21\%$} & $3.19\%$ & $3.25\%$ & $3.48\%$ & $2.84\%$ & $2.19\%$
\\
Cross Sensor~\textcolor{calib3d_red}{$\circ$} & \textcolor{gray}{$4.25\%$} & $4.15\%$ & $4.20\%$ & $4.28\%$ & $4.06\%$ & $3.20\%$ & \textcolor{gray}{$3.15\%$} & $3.13\%$ & $3.18\%$ & $3.43\%$ & $3.17\%$ & $2.96\%$
\\\midrule
Average~\textcolor{calib3d_red}{$\bullet$} & \textcolor{gray}{$6.78\%$} & $6.57\%$ & $6.62\%$ & $6.67\%$ & $5.74\%$ & $5.56\%$ & \textcolor{gray}{$6.10\%$} & $6.01\%$ & $6.07\%$ & $6.29\%$ & $5.22\%$ & $5.03\%$
\\\bottomrule
\end{tabular}
}
\vspace{-0.25cm}
\end{table*}

\subsection{Domain-Shift Uncertainty}

Beyond in-domain scenarios, we also explore uncertainty estimates under more challenging domain-shift conditions. Following the out-of-domain (OoD) settings from \textit{Robo3D} \cite{kong2023robo3D}, we train 3D scene understanding models on in-domain data and test them under OoD conditions.

\noindent\textbf{Common Corruptions.}
Real-world 3D data often include inherent measurement noise and variations. The first four rows of \cref{tab:robo3d} illustrate these issues, showing that models experience significantly higher calibration errors under corruptions caused by adverse weather conditions, including \textit{fog}, \textit{wet ground}, and \textit{snow}. The degradation from \textit{motion blur} further emphasizes the importance of network calibration for reliable 3D scene understanding.

\begin{figure}[t]
    \centering
    \begin{subfigure}[b]{0.312\linewidth}
        \centering
        \includegraphics[width=\linewidth]{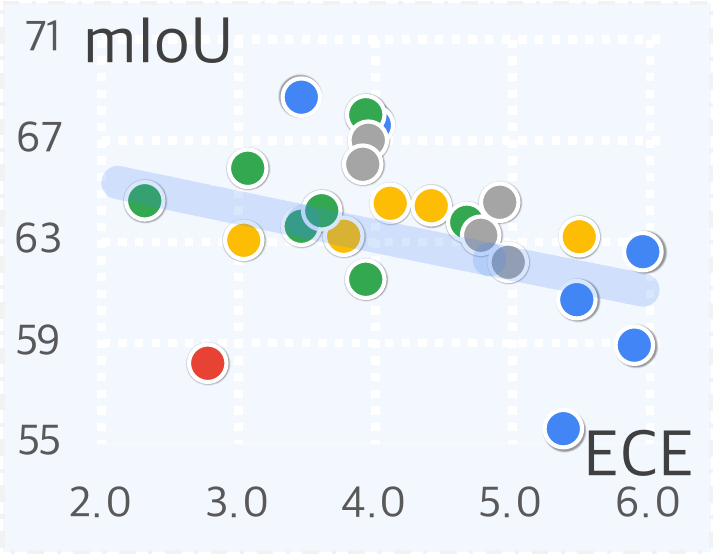}
        \caption{ECE \textit{vs.} mIoU}
        \label{fig:miou}
    \end{subfigure}~
    \begin{subfigure}[b]{0.312\linewidth}
        \centering
        \includegraphics[width=\linewidth]{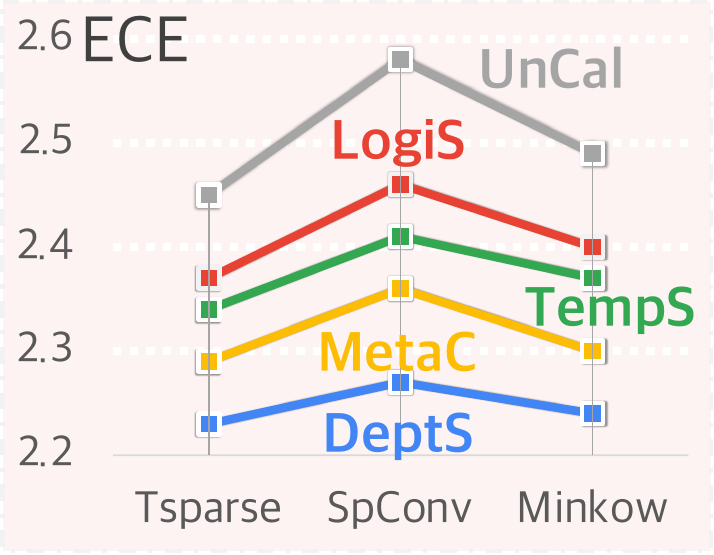}
        \caption{Backend}
        \label{fig:backend}
    \end{subfigure}~
    \begin{subfigure}[b]{0.312\linewidth}
        \centering
        \includegraphics[width=\linewidth]{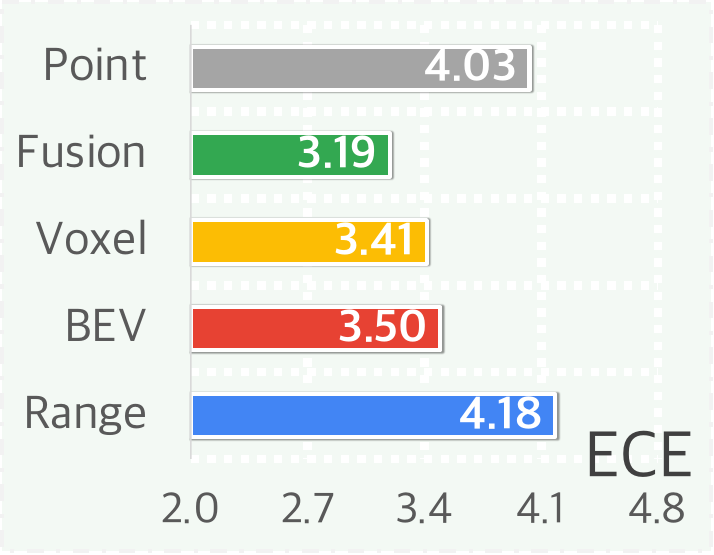}
        \caption{LiDAR Modality}
        \label{fig:modality}
    \end{subfigure}
    \vspace{-0.2cm}
    \caption{Ablation studies on (a) relationships between calibration error and intersection-over-union scores, (b) calibration errors of MinkUNet \cite{choy2019minkowski} using different sparse convolution backends, and (c) average calibration errors of different LiDAR representations.}
    \label{fig:test}
    \vspace{-0.3cm}
\end{figure}

\noindent\textbf{Sensor Failures.}
The \textit{beam missing}, \textit{crosstalk}, \textit{incomplete echo}, and \textit{cross sensor} scenarios in \cref{tab:robo3d} expose the vulnerability of existing 3D scene understanding models to various sensor failures. Compared to prior calibration methods \cite{guo2017calib,kull2019dirichlet,wang2023selective,ma2021metaCal}, DeptS is more stable in providing uncertainty estimates under these OoD conditions. This robustness is essential for achieving more reliable 3D scene understanding, particularly in safety-critical applications.

\subsection{Ablation Study}

In this section, we study several key settings that coped closely with current 3D scene understanding research. To control variables, unless otherwise specified, we use a MinkUNet-18 \cite{choy2019minkowski} model with voxel size of $0.10$ m$^3$, common augmentations, and TorchSparse \cite{tang2022torchsparse,tang2023torchsparse++} backend on the \textit{nuScenes} \cite{fong2022panoptic-nuScenes} dataset throughout this ablation study.

\begin{table}[t]
\centering
\caption{Ablation study on the uncertainty of 3D segmentation networks with different model capacities (\# of parameters).}
\vspace{-0.2cm}
\scalebox{0.75}{
\begin{tabular}{p{2cm}<{\raggedleft}|p{1.1cm}<{\centering}p{1.1cm}<{\centering}p{1.1cm}<{\centering}|p{1.25cm}<{\centering}|p{1.25cm}<{\centering}}
    \toprule
    \textbf{MinkUNet} & \textcolor{gray}{UnCal} & TempS & MetaC & \cellcolor{calib3d_blue!10}\textcolor{calib3d_blue}{DeptS} & \textbf{mIoU}
    \\\midrule\midrule
    $14\times$Layer~\textcolor{calib3d_blue}{$\bullet$} & \textcolor{gray}{$2.25\%$} & $2.21\%$ & $2.19\%$ & $2.08\%$ & $73.48\%$
    \\
    $18\times$Layer~\textcolor{calib3d_red}{$\bullet$} & \textcolor{gray}{$2.45\%$} & $2.34\%$ & $2.29\%$ & $2.23\%$ & $76.19\%$
    \\
    $34\times$Layer~\textcolor{calib3d_yellow}{$\bullet$} & \textcolor{gray}{$2.50\%$} & $2.38\%$ & $2.32\%$ & $2.22\%$ & $76.99\%$
    \\
    $50\times$Layer~\textcolor{calib3d_green}{$\bullet$} & \textcolor{gray}{$2.56\%$} & $2.41\%$ & $2.39\%$ & $2.30\%$ & $77.70\%$
    \\
    $101\times$Layer~\textcolor{calib3d_gray}{$\bullet$} & \textcolor{gray}{$2.60\%$} & $2.46\%$ & $2.35\%$ & $2.20\%$ & $79.69\%$
    \\\bottomrule
\end{tabular}}
\label{tab:networks}
\vspace{-0.1cm}
\end{table}

\begin{table}[t]
\centering
\caption{Ablation study on the uncertainty of 3D segmentation networks with different 3D data augmentation methods.}
\vspace{-0.2cm}
\scalebox{0.75}{
\begin{tabular}{p{2.3cm}<{\raggedleft}|p{1.1cm}<{\centering}p{1.1cm}<{\centering}p{1.1cm}<{\centering}|p{1.25cm}<{\centering}|p{1.25cm}<{\centering}}
    \toprule
    \textbf{Augment} & \textcolor{gray}{UnCal} & TempS & MetaC & \cellcolor{calib3d_blue!10}\textcolor{calib3d_blue}{DeptS} & \textbf{mIoU}
    \\\midrule\midrule
    Common~\textcolor{calib3d_blue}{$\bullet$} & \textcolor{gray}{$2.45\%$} & $2.34\%$ & $2.29\%$ & $2.23\%$ & $76.19\%$
    \\
    PolarMix~\textcolor{calib3d_red}{$\bullet$} & \textcolor{gray}{$2.39\%$} & $2.35\%$ & $2.30\%$ & $2.20\%$ & $76.19\%$
    \\
    LaserMix~\textcolor{calib3d_yellow}{$\bullet$} & \textcolor{gray}{$2.22\%$} & $2.21\%$ & $2.18\%$ & $2.15\%$ & $76.39\%$
    \\
    FrustumMix~\textcolor{calib3d_green}{$\bullet$} & \textcolor{gray}{$2.27\%$} & $2.26\%$ & $2.25\%$ & $2.21\%$ & $76.43\%$
    \\
    Combo~\textcolor{calib3d_gray}{$\bullet$} & \textcolor{gray}{$2.21\%$} & $2.21\%$ & $2.23\%$ & $2.18\%$ & $77.15\%$
    \\\bottomrule
\end{tabular}}
\label{tab:augment}
\vspace{-0.1cm}
\end{table}

\noindent\textbf{Network Capacity.}
Prior studies \cite{guo2017calib,wang2023selective,ma2021metaCal} have shown that larger 2D models tend to be less calibrated than smaller ones. From \cref{tab:networks}, we observed a similar trend in 3D scene understanding models. Models with fewer parameters exhibit lower calibration errors, albeit being less accurate. This raises concerns about the development of large 3D models for safety-critical applications. Special attention should be drawn when designing models with larger capacities since they are prone to be less calibrated in practice.

\noindent\textbf{3D Data Augmentations.}
Recent advancements in 3D data augmentations have exhibited superior 3D segmentation accuracy. In \cref{tab:augment}, we benchmark popular techniques, namely LaserMix \cite{kong2022lasermix}, PolarMix \cite{xiao2022polarmix}, and FrustumMix \cite{xu2023frnet}, on their efficacy in uncertainty estimation. We observe large improvements in them in delivering reliable uncertain estimates, compared to their baselines.

\begin{table}[t]
\centering
\caption{Ablation study on the uncertainty of CENet \cite{cheng2022cenet} with different \# of range view cells on SemanticKITTI \cite{behley2019semanticKITTI}.}
\vspace{-0.2cm}
\scalebox{0.75}{
\begin{tabular}{p{2cm}<{\raggedleft}|p{1.1cm}<{\centering}p{1.1cm}<{\centering}p{1.1cm}<{\centering}|p{1.25cm}<{\centering}|p{1.25cm}<{\centering}}
    \toprule
    \textbf{\# of Cells} & \textcolor{gray}{UnCal} & TempS & MetaC & \cellcolor{calib3d_blue!10}\textcolor{calib3d_blue}{DeptS} & \textbf{mIoU}
    \\\midrule\midrule
    $64\times512$~\textcolor{calib3d_blue}{$\bullet$} & \textcolor{gray}{$5.65\%$} & $4.01\%$ & $3.16\%$ & $3.09\%$ & $60.92\%$
    \\
    $64\times1024$~\textcolor{calib3d_red}{$\bullet$} & \textcolor{gray}{$5.88\%$} & $4.04\%$ & $3.24\%$ & $3.16\%$ & $62.04\%$
    \\
    $64\times2048$~\textcolor{calib3d_yellow}{$\bullet$} & \textcolor{gray}{$5.95\%$} & $3.93\%$ & $3.21\%$ & $3.10\%$ & $61.18\%$
    \\
    $64\times3072$~\textcolor{calib3d_green}{$\bullet$} & \textcolor{gray}{$6.00\%$} & $3.45\%$ & $2.85\%$ & $2.71\%$ & $60.66\%$
    \\
    $64\times4096$~\textcolor{calib3d_gray}{$\bullet$} & \textcolor{gray}{$6.21\%$} & $3.19\%$ & $2.90\%$ & $2.73\%$ & $58.68\%$
    \\\bottomrule
\end{tabular}}
\label{tab:range_cells}
\vspace{-0.1cm}
\end{table}

\begin{table}[t]
\centering
\caption{Ablation study on the uncertainty of MinkUNet-18 \cite{choy2019minkowski} with different voxel sizes (cubic shape) on nuScenes \cite{fong2022panoptic-nuScenes}.}
\vspace{-0.2cm}
\scalebox{0.75}{
\begin{tabular}{p{2.3cm}<{\raggedleft}|p{1.1cm}<{\centering}p{1.1cm}<{\centering}p{1.1cm}<{\centering}|p{1.25cm}<{\centering}|p{1.25cm}<{\centering}}
    \toprule
    \textbf{Voxel Size} & \textcolor{gray}{UnCal} & TempS & MetaC & \cellcolor{calib3d_blue!10}\textcolor{calib3d_blue}{DeptS} & \textbf{mIoU}
    \\\midrule\midrule
    $0.05$ meter$^3$~\textcolor{calib3d_blue}{$\bullet$} & \textcolor{gray}{$2.32\%$} & $2.30\%$ & $2.28\%$ & $2.23\%$ & $71.59\%$
    \\
    $0.07$ meter$^3$~\textcolor{calib3d_red}{$\bullet$} & \textcolor{gray}{$2.34\%$} & $2.28\%$ & $2.27\%$ & $2.21\%$ & $75.14\%$
    \\
    $0.10$ meter$^3$~\textcolor{calib3d_yellow}{$\bullet$} & \textcolor{gray}{$2.45\%$} & $2.34\%$ & $2.29\%$ & $2.23\%$ & $76.19\%$
    \\
    $0.15$ meter$^3$~\textcolor{calib3d_green}{$\bullet$} & \textcolor{gray}{$2.48\%$} & $2.43\%$ & $2.28\%$ & $2.21\%$ & $75.92\%$
    \\
    $0.20$ meter$^3$~\textcolor{calib3d_gray}{$\bullet$} & \textcolor{gray}{$2.68\%$} & $2.60\%$ & $2.36\%$ & $2.25\%$ & $75.53\%$
    \\\bottomrule
\end{tabular}}
\label{tab:voxel_size}
\vspace{-0.2cm}
\end{table}

\noindent\textbf{3D Rasterization.}
The resolution of 3D rasterization impacts both accuracy and calibration error. As seen in \cref{tab:range_cells} and \cref{tab:voxel_size}, Optimal segmentation accuracy typically occurs at moderate resolutions. However, calibration error poses a clear correlation with the 3D rasterization, where more range view cells or smaller voxel sizes lead to increased calibration errors and vice versa. Careful consideration is required when configuring resolutions for training and evaluation across different LiDAR representations.

\noindent\textbf{Segmentation Accuracy.}
We find a distinct correlation between calibration errors and 3D segmentation accuracy, \ie, mIoU scores, as shown in \cref{fig:miou}. Similar to the observation drawn in \cite{vaze2021openset}, we find that a model with higher task accuracy is likely to have a relatively lower calibration error.

\noindent\textbf{SparseConv Backends.}
We compare the behaviors of MinkUNet \cite{choy2019minkowski} trained using different sparse convolution backends, \ie,
MinkowskiEngine \cite{choy2019minkowski}, SpConv \cite{yan2018second,spconv2022}, and TorchSparse \cite{tang2022torchsparse,tang2023torchsparse++}, and display the results in \cref{fig:backend}. In a general sense, SpConv \cite{yan2018second,spconv2022} tends to yield a higher calibration error than the other two backends. Our DeptS shows better performance across three scenarios.

\noindent\textbf{3D Representations.}
In \cref{fig:modality}, we calculate the average calibration errors from all models benchmarked in \cref{tab:benchmark} and split them into groups based on the use of 3D representations. As can be seen, models with point and range view representations are less calibrated than other modalities. Fusion-based models exhibit superiority in general, which showcases their efficacy in real-world cases.

\vspace{0.12cm}
\section{Conclusion}
\label{sec:conclusion}

% In this study, we have presented \textbf{Calib3D}, a pioneering benchmark aimed at evaluating the reliability and accuracy of uncertainty estimates of 3D scene understanding models. Through our comprehensive assessment of state-of-the-art models across diverse 3D datasets, we have illuminated the critical challenges these models face in delivering confident and accurate predictions, especially within the context of safety-critical applications. Our findings reveal a significant gap in the calibration capabilities of current 3D scene understanding models, which often exhibit commendable accuracy but fall short in aligning their confidence levels with their predictive accuracy.
% Addressing this crucial gap, we introduced \textbf{DeptS}, a depth-aware scaling method specifically tailored to enhance the calibration of 3D models. By dynamically adjusting the logits distribution based on depth-correlated temperature scaling, the calibration performance is significantly improved. We hope that Calib3D and DeptS will inspire further research and innovation in the field of reliable 3D scene understanding.

We introduced \textbf{Calib3D}, a benchmark that focuses on evaluating the reliability of uncertainty estimates in 3D scene understanding models. Through extensive evaluations of state-of-the-art models across diverse 3D datasets, we highlighted critical challenges in delivering confident and accurate predictions, particularly in safety-critical applications. Our results expose a significant gap in the calibration of current 3D models, which often achieve high accuracy but struggle to align confidence with predictive accuracy. To address this, we proposed \textbf{DeptS}, a depth-aware scaling method that enhances calibration by adjusting logits based on depth-correlated temperature scaling. We hope that Calib3D and DeptS will inspire further research and innovation in the field of reliable 3D scene understanding.

\vspace{0.1cm}
{\small\noindent\textbf{Acknowledgments.}
This work is supported by the Ministry of Education, Singapore, under MOE AcRF Tier 2 (MOET2EP20221-0012), NTU NAP, and RIE2020 Industry Alignment Fund -- Industry Collaboration Projects (IAF-ICP) Funding Initiative, as well as cash and in-kind contribution from the industry partner(s).}

%%%%%%%%% APPENDIX
\section*{Appendix}
\startcontents[appendices]
\printcontents[appendices]{l}{1}{\setcounter{tocdepth}{3}}

\section{Calib3D Benchmark}
\label{sec:supp_calib3d}

In this section, we elaborate on additional details about the proposed Calib3D benchmark, including basic configurations regarding the datasets (Sec.~\ref{subsec:3d_datasets}), models (Sec.~\ref{subsec:3d_models}), evaluation protocols (Sec.~\ref{subsec:protocol}), and license (Sec.~\ref{subsec:license}).

\subsection{3D Datasets}
\label{subsec:3d_datasets}
The Calib3D benchmark encompasses a total of \textbf{10} popular datasets in the area of 3D scene understanding, with a diverse spectrum of dataset configurations regarding data collections, label mappings, and annotation protocols. \cref{tab:dataset_summary} provides an overview of the datasets used in our benchmark. The key features of each dataset are summarized as follows.

\begin{table*}[t]
\caption{Summary of 3D datasets encompassed in the \textbf{Calib3D} benchmark. A total of \textbf{ten} 3D datasets have been used in our benchmark, including $^1$\textit{nuScenes} \cite{fong2022panoptic-nuScenes}, $^2$\textit{SemanticKITTI} \cite{behley2019semanticKITTI}, $^3$\textit{Waymo Open} \cite{sun2020waymoOpen}, $^4$\textit{SemanticPOSS} \cite{pan2020semanticPOSS}, $^5$\textit{SemanticSTF} \cite{xiao2023semanticSTF}, $^6$\textit{ScribbleKITTI} \cite{unal2022scribbleKITTI}, $^7$\textit{Synth4D} \cite{saltori2020synth4D}, $^8$\textit{S3DIS} \cite{armeni2016s3dis}, and $^9$\textit{nuScenes-C} and $^{10}$\textit{SemanticKITTI-C} from the Robo3D benchmark \cite{kong2023robo3D}. Each dataset sheds light on a specific data acquisition and annotation protocol, such as different LiDAR sensor setups, adverse weather conditions, weak annotations, synthetic data, indoor scenes, and out-of-domain corruptions. The images shown here are adopted from the original dataset papers.}
\vspace{-0.1cm}
\centering
\resizebox{\textwidth}{!}{
\begin{tabular}{c|c|c|c|c}
\toprule
    nuScenes & SemanticKITTI & Waymo Open & SemanticPOSS & SemanticSTF
\\\midrule
    \begin{minipage}[b]{0.45\columnwidth}\centering\raisebox{-.3\height}{\includegraphics[width=\linewidth]{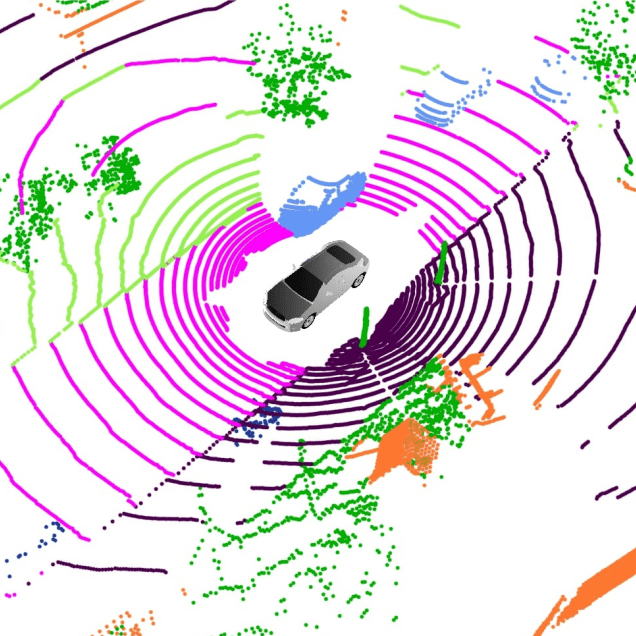}}\end{minipage} & 
    \begin{minipage}[b]{0.45\columnwidth}\centering\raisebox{-.3\height}{\includegraphics[width=\linewidth]{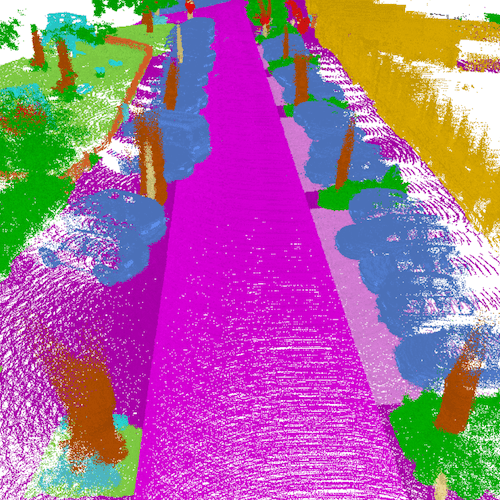}}\end{minipage} & 
    \begin{minipage}[b]{0.45\columnwidth}\centering\raisebox{-.3\height}{\includegraphics[width=\linewidth]{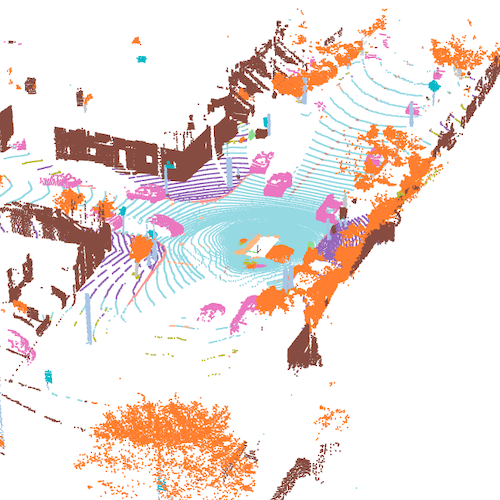}}\end{minipage} & 
    \begin{minipage}[b]{0.45\columnwidth}\centering\raisebox{-.3\height}{\includegraphics[width=\linewidth]{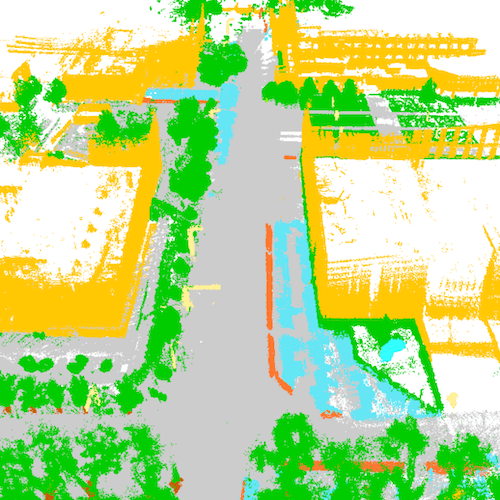}}\end{minipage} & 
    \begin{minipage}[b]{0.45\columnwidth}\centering\raisebox{-.3\height}{\includegraphics[width=\linewidth]{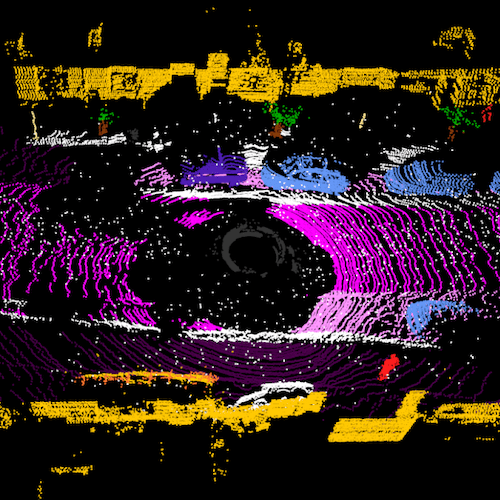}}\end{minipage}
\\\midrule
    ScribbleKITTI & Synth4D & S3DIS & nuScenes-C & SemanticKITTI-C
\\\midrule
    \begin{minipage}[b]{0.45\columnwidth}\centering\raisebox{-.3\height}{\includegraphics[width=\linewidth]{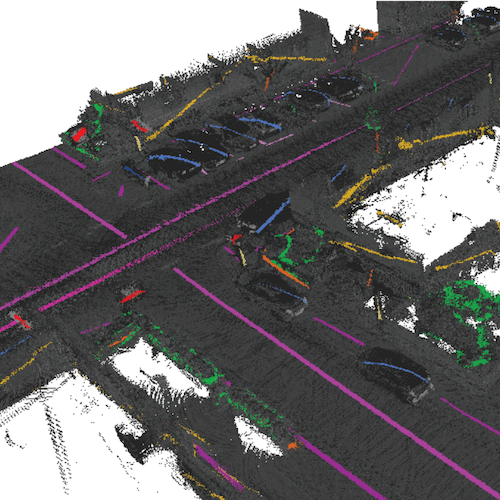}}\end{minipage} & \begin{minipage}[b]{0.45\columnwidth}\centering\raisebox{-.3\height}{\includegraphics[width=\linewidth]{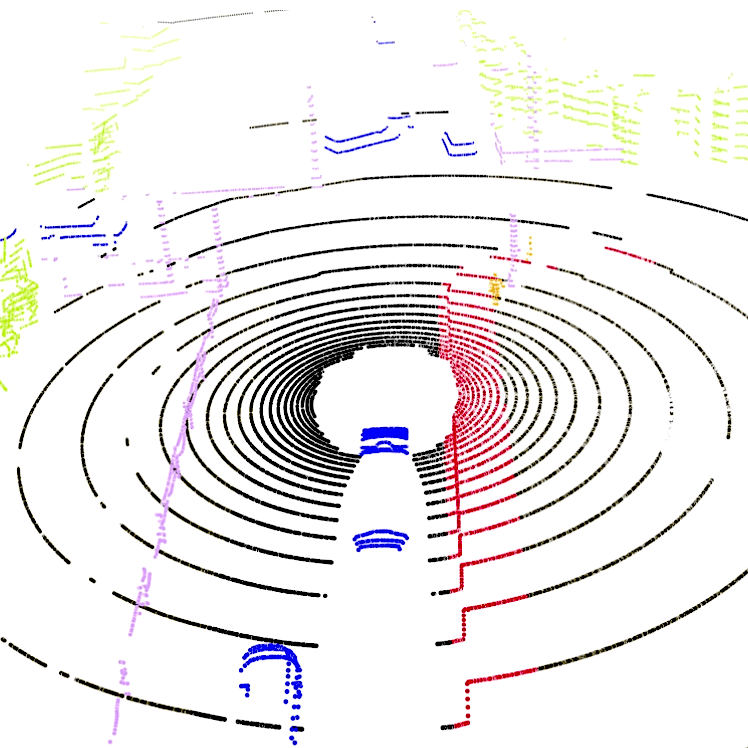}}\end{minipage} & 
    \begin{minipage}[b]{0.45\columnwidth}\centering\raisebox{-.3\height}{\includegraphics[width=\linewidth]{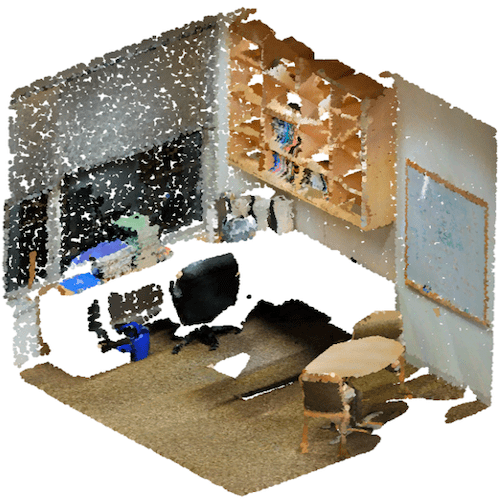}}\end{minipage} & 
    \begin{minipage}[b]{0.45\columnwidth}\centering\raisebox{-.3\height}{\includegraphics[width=\linewidth]{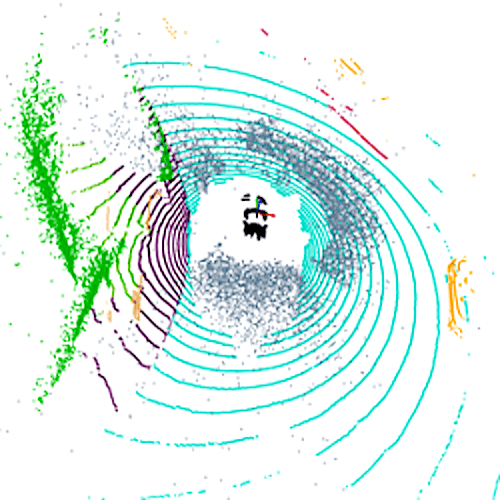}}\end{minipage} & 
    \begin{minipage}[b]{0.45\columnwidth}\centering\raisebox{-.3\height}{\includegraphics[width=\linewidth]{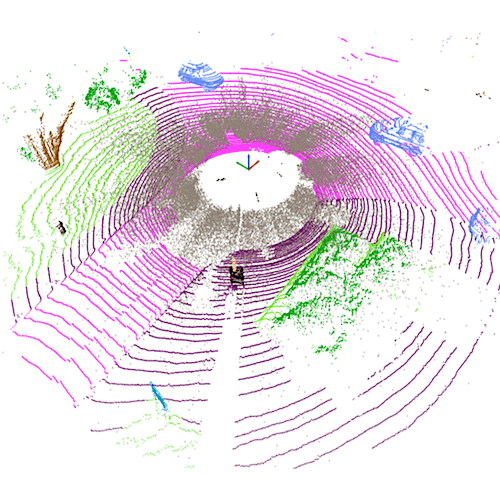}}\end{minipage}

\\\bottomrule
\end{tabular}}
\vspace{0.2cm}
\label{tab:dataset_summary}
\end{table*}

\begin{itemize}
    \item \textbf{nuScenes} \cite{fong2022panoptic-nuScenes} is one of the most popular driving datasets in autonomous vehicle research, featuring multimodal data from Boston and Singapore. It contains $1000$ scenes with $1.1$ billion annotated LiDAR points acquired by a Velodyne HDL32E LiDAR sensor. In this work, we use the \textit{lidarseg} subset of the Panoptic-nuScenes dataset, which provides point-wise class and instance labels across 16 merged semantic categories. For more information: \url{https://www.nuscenes.org/nuscenes}.
    
    \item \textbf{SemanticKITTI} \cite{behley2019semanticKITTI} offers $22$ densely labeled LiDAR sequences of urban street scenes, making it one of the most prevailing benchmarks for LiDAR-based semantic scene understanding. The point clouds are acquired by a Velodyne HDL-64E LiDAR sensor and are annotated with a total of 19 semantic categories. For more information: \url{http://semantic-kitti.org}.

    \item \textbf{Waymo Open Dataset (WOD)} \cite{sun2020waymoOpen} is a large-scale dataset for autonomous driving. The 3D semantic segmentation subset of WOD comprises $1150$ scenes, which are further split into $798$ training, $202$ validation, and $150$ testing scenes, corresponding to $23691$ training scans, $5976$ validation scans, and $2982$ testing scans, respectively. The LiDAR scans are annotated across $22$ semantic categories. For more information: \url{https://waymo.com/open}.
    
    \item \textbf{SemanticPOSS} \cite{pan2020semanticPOSS} is constructed with a special focus on dynamic scenes. It includes $2988$ scans from a 40-beam Hesai Pandora LiDAR sensor, offering insights into scene dynamics at Peking University’s campus. For more information: \url{https://www.poss.pku.edu.cn/semanticposs}.

    \item \textbf{SemanticSTF} \cite{xiao2023semanticSTF} is built on the STF dataset \cite{bijelic2020stf}. It features $2076$ scans under various weather conditions in the real world, serving as a testbed for assessing model robustness. The point clouds are acquired by a Velodyne HDL64 S3D LiDAR sensor under snowy, foggy, and rainy scenarios. For more information:  \url{https://github.com/xiaoaoran/SemanticSTF}.

    \item \textbf{ScribbleKITTI} \cite{unal2022scribbleKITTI} is an extension of the SemanticKITTI \cite{behley2019semanticKITTI} dataset. It introduces weakly-supervised annotations through line scribbles, offering a cost-effective labeling approach for $19130$ LiDAR scans, which are under the same data splits and semantic annotations of cite{behley2019semanticKITTI}. For more information: \url{https://github.com/ouenal/scribblekitti}.

    \item \textbf{Synth4D} \cite{saltori2020synth4D} was collected utilizing CARLA simulations \cite{dosovitskiy2017carla}. The Synth4D-nuScenes subset contains about $20000$ labeled point clouds for testing model performance in virtual urban and rural scenes, where the label mappings are aligned with that of the nuScenes \cite{fong2022panoptic-nuScenes} dataset. For more information: \url{https://github.com/saltoricristiano/gipso-sfouda}.

    \item \textbf{S3DIS} \cite{armeni2016s3dis} is a comprehensive collection of point clouds for indoor spaces. It encompasses detailed scans from six large-scale indoor areas that include over 215 million points and covers more than 6,000 square meters. Each point in the dataset is annotated with one of several semantic labels corresponding to different object categories like walls, floors, chairs, tables, \etc. For more information: \url{http://buildingparser.stanford.edu/dataset.html}.

    \item \textbf{nuScenes-C} \cite{kong2023robo3D} is part of the 3D robustness benchmarks in Robo3D \cite{kong2023robo3D} and is built based on the nuScenes \cite{fong2022panoptic-nuScenes} dataset. It focuses on the 3D model's out-of-distribution robustness against eight types of common corruptions, offering a platform for testing under diverse adverse conditions. For more information: \url{https://github.com/ldkong1205/Robo3D}.

    \item \textbf{SemanticKITTI-C} \cite{kong2023robo3D} shares the same common corruption types with nuScenes-C and is built based on the SemanticKITTI \cite{behley2019semanticKITTI} dataset. For more information: \url{https://github.com/ldkong1205/Robo3D}.
    
\end{itemize}

\subsection{3D Models}
\label{subsec:3d_models}
The Calib3D benchmark encompasses a total of \textbf{28} state-of-the-art models in the area of 3D scene understanding, with a diverse spectrum of LiDAR representations, network architectures, and pre-/post-processing. \cref{tab:3d_models} provides a summary of the models used, including their LiDAR modalities and key features.

\begin{table*}[t]
\centering
\caption{Summary of 3D models encompassed in the \textbf{Calib3D} benchmark. We categorize models into five distinct groups, based on their LiDAR representations, \ie, $^1$\textit{range view}, $^2$\textit{bird's eye view}, $^3$\textit{sparse voxel}, $^4$\textit{multi-view fusion}, and $^5$\textit{raw point}. Each model sheds light on a specific network structure and model configuration.
}
\vspace{-0.1cm}
\label{tab:3d_models}
\scalebox{0.9}{
\begin{tabular}{l|l|l|l}
\toprule
\textbf{Model} & \textbf{Modality} & \textbf{Key Feature} & \textbf{Ref}
\\\midrule\midrule
RangeNet$^{++}$ & \textcolor{calib3d_blue}{$\bullet$}~Range View & The first range view LiDAR segmentation model with a FCN structure & \cite{milioto2019rangenet++}
\\
SalsaNext  & \textcolor{calib3d_blue}{$\bullet$}~Range View & Uncertainty-aware range view segmentation with dilation modules & \cite{cortinhal2020salsanext}
\\
FIDNet & \textcolor{calib3d_blue}{$\bullet$}~Range View & Fully interpolation encoding for better range view post processing & \cite{zhao2021fidnet}
\\
CENet & \textcolor{calib3d_blue}{$\bullet$}~Range View & Concise and efficient range view learning with unified model structure & \cite{cheng2022cenet}
\\
RangeViT & \textcolor{calib3d_blue}{$\bullet$}~Range View & Replace ResNet backbone with ViT for enhancing range view learning  & \cite{ando2023rangevit}
\\
RangeFormer & \textcolor{calib3d_blue}{$\bullet$}~Range View & Combine RangeAug, RangePost, and RangeFormer for better results & \cite{kong2023rethinking}
\\
FRNet & \textcolor{calib3d_blue}{$\bullet$}~Range View & Frustum-range fusion \& interpolation for scalable LiDAR segmentation  & \cite{xu2023frnet}
\\\midrule
PolarNet & \textcolor{calib3d_red}{$\bullet$}~Bird's Eye View & Point cloud embedding using polar coordinates for real-time processing & \cite{zhou2020polarNet}
\\\midrule
MinkUNet$_{18}$ & \textcolor{calib3d_yellow}{$\bullet$}~Sparse Voxel & Highly efficient sparse convolution operators with cubic voxel grids & \cite{choy2019minkowski}
\\
MinkUNet$_{34}$ & \textcolor{calib3d_yellow}{$\bullet$}~Sparse Voxel & Enhanced MinkUNet structure with a larger segmentation backbone & \cite{choy2019minkowski}
\\
Cylinder3D & \textcolor{calib3d_yellow}{$\bullet$}~Sparse Voxel & Cylindrical voxel representation for balanced LiDAR points encoding & \cite{zhu2021cylindrical}
\\
SpUNet$_{18}$ & \textcolor{calib3d_yellow}{$\bullet$}~Sparse Voxel & MinkUNet structure with SpConv operators for efficient 3D learning & \cite{spconv2022}
\\
SpUNet$_{34}$ & \textcolor{calib3d_yellow}{$\bullet$}~Sparse Voxel & Enhanced SpUNet structure with a larger segmentation backbone & \cite{spconv2022}
\\\midrule
RPVNet & \textcolor{calib3d_green}{$\bullet$}~Multi-View Fusion & Multi-view fusion of range, point, and voxel for modality interactions & \cite{xu2021rpvnet}
\\
2DPASS & \textcolor{calib3d_green}{$\bullet$}~Multi-View Fusion & Distillation from images to enhance point cloud feature learning & \cite{yan2022dpass}
\\
SPVCNN$_{18}$ & \textcolor{calib3d_green}{$\bullet$}~Multi-View Fusion & Efficient sparse point-voxel convolutions \& a lightweight architecture & \cite{tang2020searching} 
\\
SPVCNN$_{34}$ & \textcolor{calib3d_green}{$\bullet$}~Multi-View Fusion & Enhanced SPVCNN structure with a larger segmentation backbone & \cite{tang2020searching}
\\
CPGNet & \textcolor{calib3d_green}{$\bullet$}~Multi-View Fusion & Cascade point-grid fusion \& transformation consistency regularization & \cite{li2022cpgnet}
\\
GFNet & \textcolor{calib3d_green}{$\bullet$}~Multi-View Fusion & Complementary geometric flow fusion of range and bird's eye views & \cite{qiu2022GFNet}
\\
UniSeg & \textcolor{calib3d_green}{$\bullet$}~Multi-View Fusion & Unified multi-view representation learning and cross-view distillation & \cite{liu2023uniseg}
\\\midrule
KPConv & \textcolor{calib3d_gray}{$\bullet$}~Raw Point Input & Deformable convolutions for adaptive kernel-based geometry learning &  \cite{thomas2019kpconv}
\\
PIDS$_{1.25\times}$ & \textcolor{calib3d_gray}{$\bullet$}~Raw Point Input & Joint point interaction-dimension search with varying point densities & \cite{zhang2023pids}
\\
PIDS$_{2.0\times}$ & \textcolor{calib3d_gray}{$\bullet$}~Raw Point Input & Enhanced PIDS structure with a larger segmentation backbone & \cite{zhang2023pids}
\\
PTv2 & \textcolor{calib3d_gray}{$\bullet$}~Raw Point Input & Grouped vector attention \& partition-based pooling using Transformers & \cite{wu2022ptv2}
\\
WaffleIron & \textcolor{calib3d_gray}{$\bullet$}~Raw Point Input & Update point features by combining multi-MLPs and dense 2D CNNs & \cite{puy23waffleiron} 
\\
PointNet$^{++}$ & \textcolor{calib3d_gray}{$\bullet$}~Raw Point Input & The first hierarchical network to direct operate on point clouds & \cite{qi2017pointnet++}
\\
DGCNN & \textcolor{calib3d_gray}{$\bullet$}~Raw Point Input & Use graph convolution to dynamically update graph in feature space & \cite{wang2019dgcnn}
\\
PAConv & \textcolor{calib3d_gray}{$\bullet$}~Raw Point Input & Dynamic kernel assembling to adjust convolutions with point positions & \cite{xu2021paconv}
\\\bottomrule
\end{tabular}}
\vspace{0.2cm}
\end{table*}

\subsection{Benchmark Protocols}
\label{subsec:protocol}
In this work, to ensure fairness in comparisons, we adopt the following protocols in model evaluations:
\begin{itemize}
    \item All 3D scene understanding models are trained on the official \textit{training} set of each 3D dataset, and evaluated on data from the official \textit{validation} set. There is no overlap between training and evaluation data.
    \item To reflect the original behavior of each 3D scene understanding model, we directly use public checkpoints whenever applicable, or re-train the model using its default configuration. The acknowledgments of public checkpoints and implementations are included in Sec.~\ref{sec:supp_acknowledge}.
    \item We notice that some models (and their public checkpoints) are enhanced using extra ``tricks'' on the validation/testing sets, such as test time augmentation, model ensembling, \etc. To ensure fairness, we re-train such models to reflect their ``clean'' performance.
\end{itemize}

\subsection{License}
\label{subsec:license}
The Calib3D benchmark is released under the \textit{CC BY-NC-SA 4.0} license\footnote{\url{https://creativecommons.org/licenses/by-nc-sa/4.0/legalcode.en}.}. For licenses regarding the codebase used in the Calib3D benchmark, kindly refer to \cref{subsec:acknowledge_codebase}. For licenses regarding the 3D datasets used in the Calib3D benchmark, kindly refer to \cref{subsec:acknowledge_datasets}. For licenses regarding the model implementations used in the Calib3D benchmark, kindly refer to \cref{subsec:acknowledge_implements}.

\section{Additional Implementation Detail}
\label{sec:supp_implement}

In this section, we provide additional implementation details to help reproduce the key results shown in this work.

\subsection{3D Model Training}
\label{subsec:3d_model_training}
Our Calib3D benchmark is constructed based on the popular MMDetection3D \cite{mmdet3d} and OpenPCSeg \cite{pcseg2023} codebase, as well as several standalone implementations that have not been integrated into MMDetection3D \cite{mmdet3d} and/or OpenPCSeg \cite{pcseg2023}. Most 3D models adopt a unified training configuration, including the number of training epochs, optimizer, and learning rate scheduler. We apply common 3D data augmentations in Cartesian space, including random flipping, rotation, scaling, and jittering. The 3D models are trained using eight GPUs with a batch size of 2. The number of epochs are set as $80$ for \textit{nuScenes} and $50$ for \textit{SemanticKITTI}, \textit{Waymo Open}, \textit{SemanticPOSS}, \textit{SemanticSTF}, \textit{ScribbleKITTI}, and \textit{Synth4D}. For \textit{S3DIS}, we follow the default setups as MMDetection3D \cite{mmdet3d}.
For additional details, please refer to the corresponding codebase.

\subsection{3D Model Evaluation}
\label{subsec:3d_model_evaluation}
We evaluate the 3D models by following the conventional evaluation setups. As mentioned in Sec.~\ref{subsec:protocol}, we do not use any extra ``tricks'' on the validation/testing sets, such as test time augmentation, model ensembling, \etc. 

\subsection{PyTorch-Style ECE Calculation}
\label{subsec:ece3d_pytorch}

To facilitate reproduction, we provide a PyTorch-style code snippet for calculating the expected calibration error (ECE) on point clouds as follows.

\begin{lstlisting}[caption={PyTorch-style code snippet for calculating ECE scores on point clouds.}, label={lst:ece3d}]
import torch
import torch.nn.functional as F

def calculate_ece(logits, labels, ignore_index, n_bins=10):
    valid_index = labels != ignore_index
    logits, labels = logits[valid_index], labels[valid_index]

    bin_bound = torch.linspace(0, 1, n_bins + 1)
    lowers, uppers = bin_bound[:-1], bin_bound[1:]

    softmaxes = F.softmax(logits, dim=1)
    confs, preds = torch.max(softmaxes, 1)
    accs = preds.eq(labels)

    ece = torch.zeros(1)
    for l, u in zip(lowers, uppers):
        in_bin = confs.gt(l.item()) * confs.le(u.item())
        prop_in_bin = in_bin.float().mean()
        if prop_in_bin.item() > 0:
            acc_in_bin = accs[in_bin].float().mean()
            avg_conf_in_bin = confs[in_bin].mean()
            ece += torch.abs(avg_conf_in_bin - acc_in_bin) * prop_in_bin
            
    return ece.item()

\end{lstlisting}

\subsection{PyTorch-Style Implementation of DeptS}
\label{subsec:depts_pytorch}

To facilitate reproduction, we provide a PyTorch-style code snippet of the proposed depth-aware scaling (DeptS) method as follows.

\begin{lstlisting}[caption={PyTorch-style code snippet of the proposed depth-aware scaling (DeptS).}, label={lst:depts}]
import numpy as np
import torch
import torch.nn as nn

class Depth_Aware_Scaling(nn.Module):

    def __init__(self, threshold):
        super(Depth_Aware_Scaling, self).__init__()
        self.T1 = nn.Parameter(torch.ones(1))
        self.T2 = nn.Parameter(torch.ones(1) * 0.9)
        self.k  = nn.Parameter(torch.ones(1) * 0.1)
        self.b  = nn.Parameter(torch.zeros(1))
        self.alpha = 0.05
        self.threshold = threshold
        self.softmax = nn.Softmax(dim=-1)

    def forward(self, logits, gt, xyz):
        if self.training:
            ind = torch.argmax(logits, axis=1) == gt
            logits_pos, gt_pos = logits[ind], gt[ind]
            logits_neg, gt_neg = logits[~ind], gt[~ind]

            depth = torch.norm(xyz, p=2, dim=1)
            depth_pos, depth_neg = depth[ind], depth[~ind]

            s = np.random.randint(int(logits_pos.shape[0] * 1 / 3)) + 1
            logits = torch.cat((
                logits_neg, logits_pos[s:int(logits_pos.shape[0] / 2) + s]
            ), 0)
            gt = torch.cat((
                gt_neg, gt_pos[s:int(logits_pos.shape[0] / 2) + s]
            ), 0)
            depth = torch.cat((
                depth_neg, depth_pos[s:int(depth_pos.shape[0] / 2) + s]
            ), 0)

            prob = self.softmax(logits)

            score = torch.sum(-prob * torch.log(prob), dim=-1)
            cond_ind = score < self.threshold

            cal_logits_1, cal_gt_1 = logits[cond_ind], gt[cond_ind]
            cal_logits_2, cal_gt_2 = logits[~cond_ind], gt[~cond_ind]

            depth_coff = self.k * depth + self.b
            T1 = self.T1 * depth_coff[cond_ind].unsqueeze(dim=-1)
            T2 = self.T2 * depth_coff[~cond_ind].unsqueeze(dim=-1)

            cal_logits_1 = cal_logits_1 / T1
            cal_logits_2 = cal_logits_2 / T2

            cal_logits = torch.cat((cal_logits_1, cal_logits_2), 0)
            cal_gt = torch.cat((cal_gt_1, cal_gt_2), 0)

        else:
            prob = self.softmax(logits)

            score = torch.sum(-prob * torch.log(prob), dim=-1)
            cond_ind = score < self.threshold
            
            scaled_logits, scaled_gt = logits[cond_ind], gt[cond_ind]
            inference_logits, inference_gt = logits[~cond_ind], gt[~cond_ind]

            depth = torch.norm(xyz, p=2, dim=1).to(logits.device)
            depth_coff = self.k * depth + self.b

            T1 = self.T1 * depth_coff[cond_ind].unsqueeze(dim=-1)
            T2 = self.T2 * depth_coff[~cond_ind].unsqueeze(dim=-1)

            scaled_logits = scaled_logits / T1
            inference_logits = inference_logits / T2

            cal_logits = torch.cat((scaled_logits, inference_logits), 0)
            cal_gt = torch.cat((scaled_gt, inference_gt), 0)
        
        return cal_logits, cal_gt

\end{lstlisting}

\section{Additional Quantitative Result}
\label{sec:supp_quantitative}

In this section, we supplement additional quantitative results to better support the findings and conclusions drawn in the main body of this paper.

\subsection{Depth Correlations in LiDAR Data}

\begin{figure}[t]
    \centering
    \includegraphics[width=1.0\linewidth]{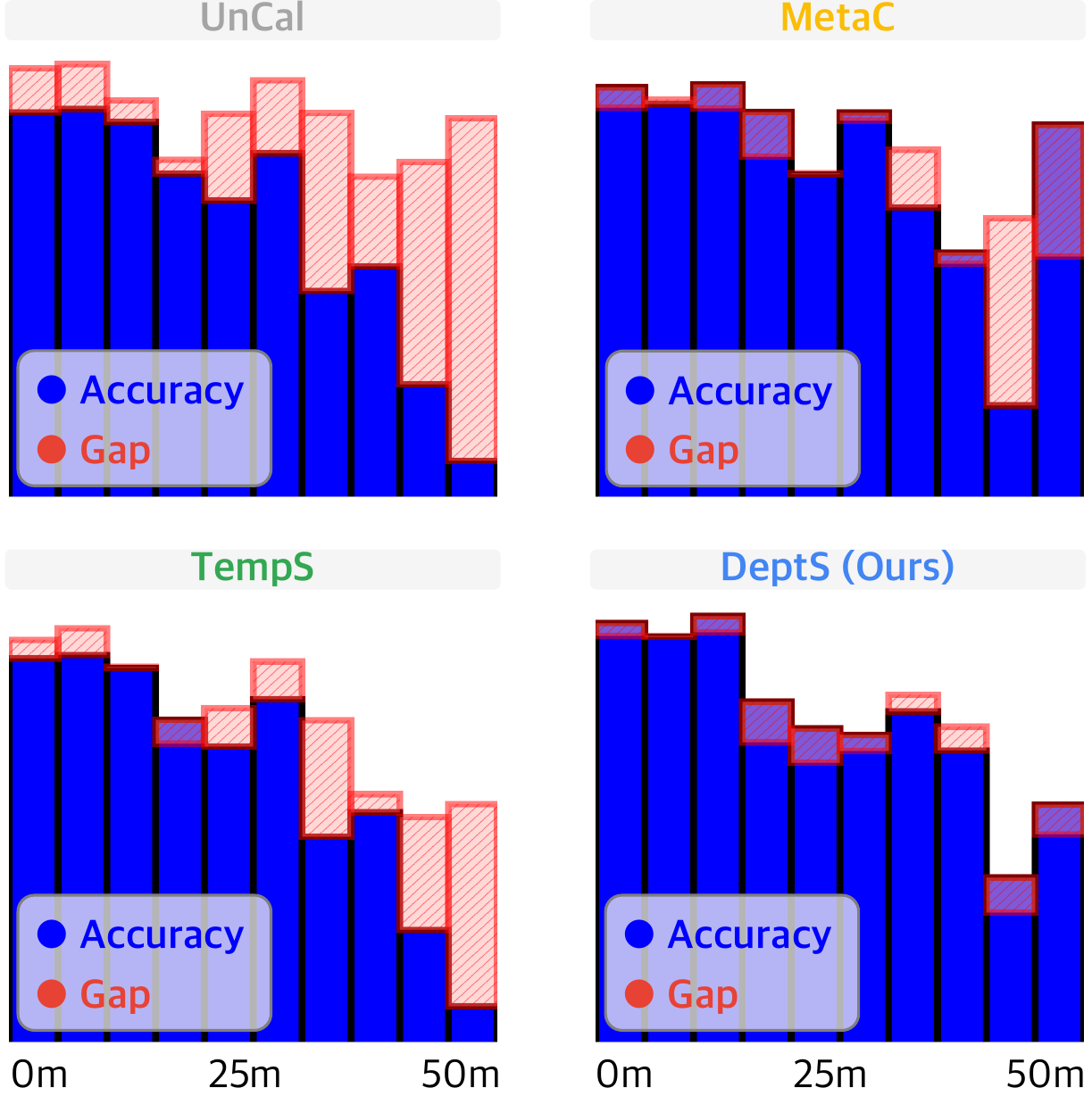}
    \vspace{-0.6cm}
    \caption{Depth-wise confidence and accuracy statistics of uncalibrated (\textcolor{gray}{UnCal}), temperature scaling (\textcolor{calib3d_green}{TempS}), meta-calibration (\textcolor{calib3d_yellow}{MetaC}), and our proposed depth-aware scaling (\textcolor{calib3d_blue}{DeptS}) methods.} 
    \label{fig:depth}
\end{figure}

As discussed in Sec.~\textcolor{calib3d_green}{3.3} of the main body of this paper, the motivation behind the depth-aware scaling method, DeptS, stems from some interesting observations from our experiments. We observe that traditional calibration techniques, effective in 2D image-based tasks, struggle with 3D data due to the unique characteristics of point clouds, such as being unordered and lacking texture. 
Through our analysis, we identified a clear correlation between calibration errors, prediction entropy, and depth. Specifically, as shown in \cref{fig:depth}, LiDAR points at greater distances from the ego vehicle often exhibit lower accuracy, yet uncalibrated models maintain high confidence in these areas, leading to substantial calibration errors \cite{kong2022lasermix,kong2024lasermix2}. This overconfidence in distant regions prompted the need for a tailored approach to address the depth-related calibration issue.

To tackle this, we propose DeptS, a method that adjusts model confidence based on depth information. By introducing a depth-correlation coefficient that reweights the temperature scaling parameters, DeptS reduces confidence for LiDAR points at larger depths, effectively mitigating the overconfidence problem. This method allows for better calibration in 3D scene understanding models, particularly in middle-to-far regions where predictions are less reliable, leading to improved calibration performance across diverse 3D datasets.

\subsection{Reliability Diagrams}
\label{subsec:reliability_diagrams}

\begin{figure*}[t]
    \begin{center}
    \includegraphics[width=\linewidth]{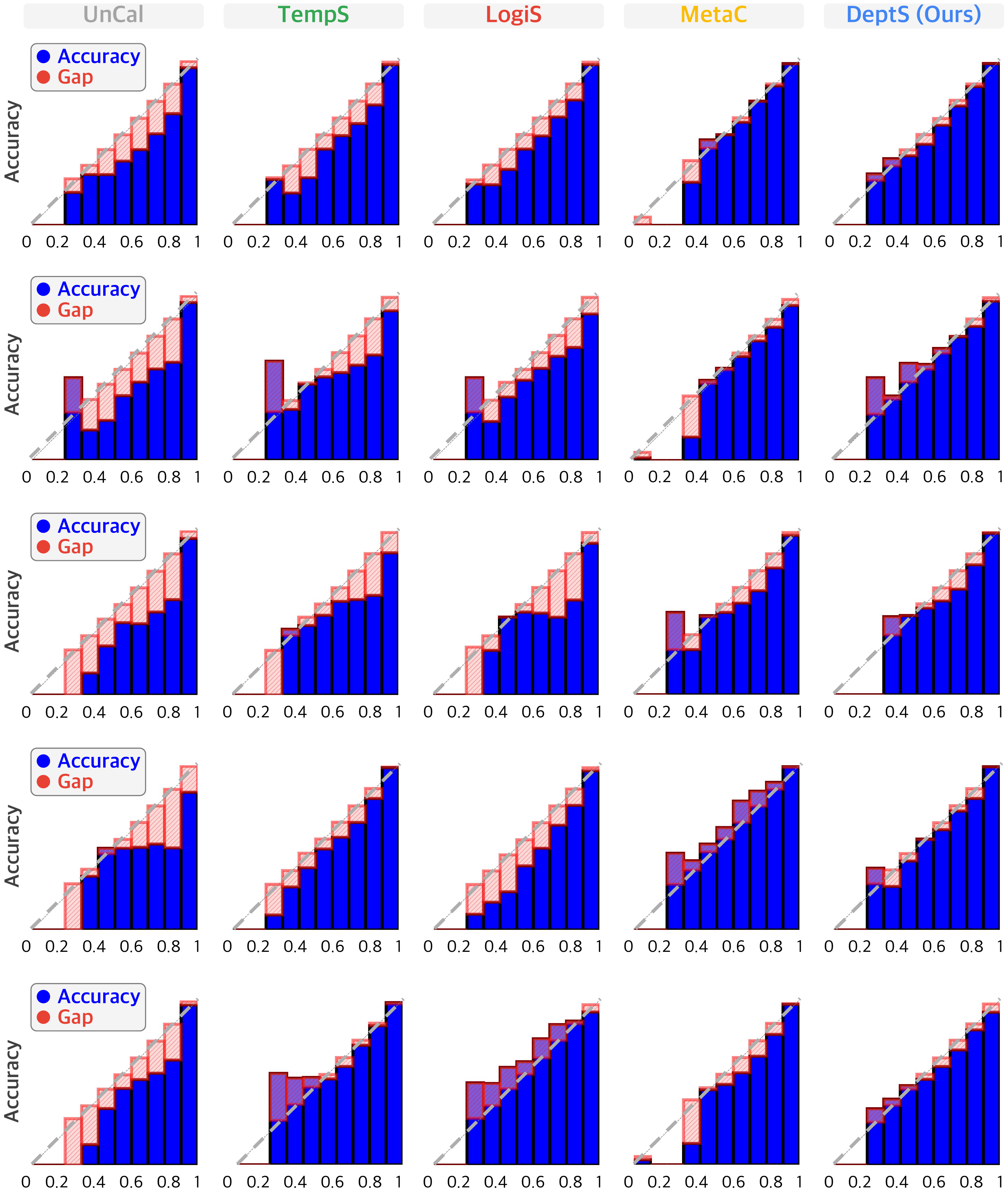}
    \end{center}
    \vspace{-0.3cm}
    \caption{The reliability diagrams of randomly sampled model predictions generated by the CENet \cite{cheng2022cenet} model on the validation set of the \textit{SemanticKITTI} \cite{behley2019semanticKITTI} dataset. \textcolor{calib3d_gray}{UnCal}, \textcolor{calib3d_green}{TempS}, \textcolor{calib3d_red}{LogiS}, \textcolor{calib3d_yellow}{MetaC}, and \textcolor{calib3d_blue}{DeptS} denote the uncalibrated, temperature, logistic, meta, and our proposed depth-aware scaling calibration methods, respectively.} 
    \label{fig:supp_bin_1}
\end{figure*}

We provide additional reliability diagrams in \cref{fig:supp_bin_1} for a more comprehensive validation of the effectiveness of our method. As can be seen, the 3D models without proper calibration (UnCal) tend to suffer from huge confidence-accuracy gaps. This inevitably leads to potential impediments to the safe operation of 3D scene understanding systems in the real world. Our compared calibration methods show effectiveness in mitigating such issues. Compared to the previous calibration methods, our DeptS exhibits superior performance across a wide spectrum of scenarios. This can be credited to the depth-aware scaling operation which encourages a more consistent prediction in depth-correlated areas.

\subsection{Domain-Shift Uncertainty Estimation}
\label{subsec:domain_shift_uncertainty}

Enhancing the uncertainty estimation capability of handling challenging scenarios is crucial for the practical usage of 3D scene understanding systems \cite{xie2024benchmarking,hao2024mapbench,li2024place3d,kong2023robo3D,kong2023robodepth,2023CLIP2Scene,chen2023towards,kong2023robodepth_challenge}.
We supplement the domain-shift uncertainty estimation results of FRNet \cite{xu2023frnet} and SPVCNN \cite{tang2020searching} in \cref{tab:robo3d_frnet} and \cref{tab:robo3d_spvcnn}, respectively. Similar to the observations drawn in the main body of this paper, we find that 3D models are vulnerable under adverse conditions. The expected calibration errors are extremely high under ``fog'', ``motion blur'', ``crosstalk'', and ``cross sensor'' corruptions, which are commonly occurring scenarios in the real world. Compared to previous calibration methods like temperature scaling and meta-calibration, our DeptS shows a more stable performance across different domain-shift scenarios. We believe such an ability will become more and more important in the future development of 3D scene understanding systems.

\begin{table}[t]
    \centering
    \caption{Comparisons between the proposed DeptS and state-of-the-art network calibration methods on the validation set of \textit{SemanticKITTI} \cite{behley2019semanticKITTI}. All ECE (the lower the better) and mIoU (the higher the better) scores reported are in percentage (\%).}
    \vspace{-0.2cm}
    \resizebox{\linewidth}{!}{
    \begin{tabular}{r|r|cc|cc}
        \toprule
        \multirow{2}{*}{\textbf{Method}} & \multirow{2}{*}{\textbf{Venue}} & \multicolumn{2}{c|}{\textbf{MinkUNet} \cite{choy2019minkowski}} & \multicolumn{2}{c}{\textbf{CENet} \cite{cheng2022cenet}}
        \\
        & & ECE & mIoU & ECE & mIoU 
        \\\midrule\midrule
        \textcolor{gray}{UnCal} & - & \textcolor{gray}{$3.04\%$} & \textcolor{gray}{$63.05\%$} & \textcolor{gray}{$5.95\%$} & \textcolor{gray}{$60.87\%$}
        \\\midrule
        TempS \cite{guo2017calib} & ICML'17 & $3.01\%$ & $63.05\%$ & $3.93\%$ & $60.87\%$
        \\
        LogiS \cite{guo2017calib} & ICML'17 & $3.08\%$ & $63.11\%$ & $3.79\%$ & $60.95\%$
        \\
        MetaC \cite{ma2021metaCal} & ICML'21 & $2.69\%$ & $62.93\%$ & $3.31\%$ & $60.81\%$
        \\
        DeepEnsemble \cite{lakshminarayanan2017deepensemble} & NeurIPS'17 & $2.99\%$ & $64.95\%$ & $5.61\%$ & $61.70\%$
        \\
        BatchEnsemble \cite{wen2020batchensemble} & ICLR'20 & $2.77\%$ & $64.70\%$ & $5.40\%$ & $62.13\%$
        \\
        MIMO \cite{havasi2021mimo} & ICLR'21 & $3.21\%$ & $63.60\%$ & $6.97\%$ & $61.62\%$
        \\
        PackedEnsemble \cite{laurent2023packedensemble} & ICLR'23 & $2.82\%$ & $63.88\%$ & $6.00\%$ & $59.81\%$
        \\\midrule
        \cellcolor{calib3d_blue!10}\textcolor{calib3d_blue}{DeptS} & \textbf{Ours} & $\mathbf{2.63\%}$ & $\mathbf{63.47\%}$ & $\mathbf{3.09\%}$ & $\mathbf{61.20\%}$
        \\\bottomrule
    \end{tabular}}
    \label{tab:recent_methods}
\end{table}

\begin{table*}[t]
\centering
\caption{The expected calibration error (ECE, the lower the better) of FRNet \cite{xu2023frnet} under eight domain-shift scenarios from \textit{nuScenes-C} and \textit{SemanticKITTI-C} in the \textit{Robo3D} benchmark \cite{kong2023robo3D}. UnCal, TempS, LogiS, DiriS, MetaC, and DeptS denote the uncalibrated, temperature, logistic, Dirichlet, meta, and our depth-aware scaling calibration methods, respectively.}
\vspace{-0.2cm}
\label{tab:robo3d_frnet}
\resizebox{\linewidth}{!}{
\begin{tabular}{r|p{1.01cm}<{\raggedleft}p{1.01cm}<{\raggedleft}p{1.01cm}<{\raggedleft}p{1.01cm}<{\raggedleft}p{1.01cm}<{\raggedleft}|p{1.01cm}<{\raggedleft}|p{1.01cm}<{\raggedleft}p{1.01cm}<{\raggedleft}p{1.01cm}<{\raggedleft}p{1.01cm}<{\raggedleft}p{1.01cm}<{\raggedleft}|p{1.01cm}<{\raggedleft}}
\toprule
\multirow{2}{*}{\textbf{Type}} & \multicolumn{6}{c|}{\textbf{nuScenes-C}} & \multicolumn{6}{c}{\textbf{SemanticKITTI-C}}
\\
& \textcolor{gray}{UnCal} & TempS & LogiS & DiriS & MetaC & \cellcolor{calib3d_blue!10}\textcolor{calib3d_blue}{DeptS} & \textcolor{gray}{UnCal} & TempS & LogiS & DiriS & MetaC & \cellcolor{calib3d_blue!10}\textcolor{calib3d_blue}{DeptS}
\\\midrule\midrule
\rowcolor{calib3d_green!8}\textcolor{calib3d_green}{\textbf{Clean}}~\textcolor{calib3d_green}{$\bullet$} & \textcolor{calib3d_green}{$2.27\%$} & \textcolor{calib3d_green}{$2.24\%$} & \textcolor{calib3d_green}{$2.22\%$} & \textcolor{calib3d_green}{$2.28\%$} & \textcolor{calib3d_green}{$2.22\%$} & \textcolor{calib3d_green}{$2.17\%$} & \textcolor{calib3d_green}{$3.46\%$} & \textcolor{calib3d_green}{$3.53\%$} & \textcolor{calib3d_green}{$3.54\%$} & \textcolor{calib3d_green}{$3.49\%$} & \textcolor{calib3d_green}{$2.83\%$} & \textcolor{calib3d_green}{$2.75\%$}
\\\midrule
Fog~\textcolor{calib3d_red}{$\circ$} & \textcolor{gray}{$3.07\%$} & $3.06\%$ & $3.07\%$ & $3.03\%$ & $3.06\%$ & $2.98\%$ & \textcolor{gray}{$13.48\%$} & $13.57\%$ & $13.66\%$ & $13.47\%$ & $12.68\%$ & $12.42\%$
\\
Wet Ground~\textcolor{calib3d_red}{$\circ$} & \textcolor{gray}{$2.46\%$} & $2.44\%$ & $2.43\%$ & $2.50\%$ & $2.56\%$ & $2.41\%$ & \textcolor{gray}{$4.01\%$} & $4.09\%$ & $4.11\%$ & $3.96\%$ & $3.32\%$ & $3.28\%$
\\
Snow~\textcolor{calib3d_red}{$\circ$} & \textcolor{gray}{$3.50\%$} & $3.42\%$ & $3.53\%$ & $3.60\%$ & $2.93\%$ & $2.78\%$ & \textcolor{gray}{$7.28\%$} & $7.39\%$ & $7.49\%$ & $7.51\%$ & $6.65\%$ & $6.63\%$
\\
Motion Blur~\textcolor{calib3d_red}{$\circ$} & \textcolor{gray}{$33.74\%$} & $33.48\%$ & $33.15\%$ & $32.15\%$ & $30.62\%$ & $28.43\%$ & \textcolor{gray}{$5.93\%$} & $6.03\%$ & $6.08\%$ & $6.55\%$ & $5.04\%$ & $4.92\%$
\\
Beam Missing~\textcolor{calib3d_red}{$\circ$} & \textcolor{gray}{$2.52\%$} & $2.51\%$ & $2.50\%$ & $2.58\%$ & $2.91\%$ & $2.48\%$ & \textcolor{gray}{$2.71\%$} & $2.71\%$ & $2.72\%$ & $2.71\%$ & $2.40\%$ & $2.36\%$
\\
Crosstalk~\textcolor{calib3d_red}{$\circ$} & \textcolor{gray}{$2.40\%$} & $2.39\%$ & $2.36\%$ & $2.38\%$ & $2.72\%$ & $2.35\%$ & \textcolor{gray}{$20.87\%$} & $21.16\%$ & $21.03\%$ & $19.84\%$ & $15.36\%$ & $14.79\%$
\\
Incomplete Echo~\textcolor{calib3d_red}{$\circ$} & \textcolor{gray}{$2.36\%$} & $2.30\%$ & $2.32\%$ & $2.34\%$ & $2.28\%$ & $2.21\%$ & \textcolor{gray}{$3.77\%$} & $3.86\%$ & $3.88\%$ & $3.82\%$ & $3.13\%$ & $3.02\%$
\\
Cross Sensor~\textcolor{calib3d_red}{$\circ$} & \textcolor{gray}{$5.24\%$} & $5.20\%$ & $5.29\%$ & $5.88\%$ & $5.34\%$ & $5.11\%$ & \textcolor{gray}{$5.08\%$} & $5.11\%$ & $5.17\%$ & $4.64\%$ & $3.91\%$ & $3.74\%$
\\\midrule
Average~\textcolor{calib3d_red}{$\bullet$} & \textcolor{gray}{$6.91\%$} & $6.85\%$ & $6.83\%$ & $6.81\%$ & $6.55\%$ & $6.09\%$ & \textcolor{gray}{$7.89\%$} & $7.99\%$ & $8.02\%$ & $7.81\%$ & $6.56\%$ & $6.40\%$
\\\bottomrule
\end{tabular}}
\vspace{0.2cm}
\end{table*}

\begin{table*}[t]
\centering
\caption{The expected calibration error (ECE, the lower the better) of SPVCNN \cite{tang2020searching} under eight domain-shift scenarios from \textit{nuScenes-C} and \textit{SemanticKITTI-C} in the \textit{Robo3D} benchmark \cite{kong2023robo3D}. UnCal, TempS, LogiS, DiriS, MetaC, and DeptS denote the uncalibrated, temperature, logistic, Dirichlet, meta, and our depth-aware scaling calibration methods, respectively.}
\vspace{-0.2cm}
\label{tab:robo3d_spvcnn}
\resizebox{\linewidth}{!}{
\begin{tabular}{r|p{1.01cm}<{\raggedleft}p{1.01cm}<{\raggedleft}p{1.01cm}<{\raggedleft}p{1.01cm}<{\raggedleft}p{1.01cm}<{\raggedleft}|p{1.01cm}<{\raggedleft}|p{1.01cm}<{\raggedleft}p{1.01cm}<{\raggedleft}p{1.01cm}<{\raggedleft}p{1.01cm}<{\raggedleft}p{1.01cm}<{\raggedleft}|p{1.01cm}<{\raggedleft}}
\toprule
\multirow{2}{*}{\textbf{Type}} & \multicolumn{6}{c|}{\textbf{nuScenes-C}} & \multicolumn{6}{c}{\textbf{SemanticKITTI-C}}
\\
& \textcolor{gray}{UnCal} & TempS & LogiS & DiriS & MetaC & \cellcolor{calib3d_blue!10}\textcolor{calib3d_blue}{DeptS} & \textcolor{gray}{UnCal} & TempS & LogiS & DiriS & MetaC & \cellcolor{calib3d_blue!10}\textcolor{calib3d_blue}{DeptS}
\\\midrule\midrule
\rowcolor{calib3d_green!8}\textcolor{calib3d_green}{\textbf{Clean}}~\textcolor{calib3d_green}{$\bullet$} & \textcolor{calib3d_green}{$2.57\%$} & \textcolor{calib3d_green}{$2.44\%$} & \textcolor{calib3d_green}{$2.49\%$} & \textcolor{calib3d_green}{$2.54\%$} & \textcolor{calib3d_green}{$2.40\%$} & \textcolor{calib3d_green}{$2.31\%$} & \textcolor{calib3d_green}{$3.46\%$} & \textcolor{calib3d_green}{$2.90\%$} & \textcolor{calib3d_green}{$3.07\%$} & \textcolor{calib3d_green}{$3.41\%$} & \textcolor{calib3d_green}{$2.36\%$} & \textcolor{calib3d_green}{$2.32\%$}
\\\midrule
Fog~\textcolor{calib3d_red}{$\circ$} & \textcolor{gray}{$8.53\%$} & $8.12\%$ & $8.23\%$ & $8.54\%$ & $7.38\%$ & $7.34\%$ & \textcolor{gray}{$13.06\%$} & $12.33\%$ & $12.57\%$ & $13.23\%$ & $11.15\%$ & $11.10\%$
\\
Wet Ground~\textcolor{calib3d_red}{$\circ$} & \textcolor{gray}{$2.80\%$} & $2.63\%$ & $2.68\%$ & $2.72\%$ & $2.63\%$ & $2.58\%$ & \textcolor{gray}{$3.52\%$} & $3.02\%$ & $3.19\%$ & $3.49\%$ & $2.76\%$ & $2.63\%$
\\
Snow~\textcolor{calib3d_red}{$\circ$} & \textcolor{gray}{$8.49\%$} & $7.76\%$ & $7.97\%$ & $8.35\%$ & $6.87\%$ & $6.61\%$ & \textcolor{gray}{$8.50\%$} & $7.70\%$ & $7.94\%$ & $8.41\%$ & $6.31\%$ & $6.26\%$
\\
Motion Blur~\textcolor{calib3d_red}{$\circ$} & \textcolor{gray}{$9.18\%$} & $8.80\%$ & $9.00\%$ & $9.33\%$ & $8.11\%$ & $7.98\%$ & \textcolor{gray}{$21.01\%$} & $19.92\%$ & $20.28\%$ & $20.41\%$ & $17.86\%$ & $17.22\%$
\\
Beam Missing~\textcolor{calib3d_red}{$\circ$} & \textcolor{gray}{$2.88\%$} & $2.70\%$ & $2.74\%$ & $2.79\%$ & $2.72\%$ & $2.67\%$ & \textcolor{gray}{$3.01\%$} & $2.64\%$ & $2.73\%$ & $3.04\%$ & $2.48\%$ & $2.45\%$
\\
Crosstalk~\textcolor{calib3d_red}{$\circ$} & \textcolor{gray}{$11.76\%$} & $11.09\%$ & $11.33\%$ & $12.01\%$ & $9.82\%$ & $9.48\%$ & \textcolor{gray}{$4.66\%$} & $4.00\%$ & $4.17\%$ & $4.49\%$ & $3.58\%$ & $3.31\%$
\\
Incomplete Echo~\textcolor{calib3d_red}{$\circ$} & \textcolor{gray}{$2.40\%$} & $2.28\%$ & $2.33\%$ & $2.39\%$ & $2.30\%$ & $2.24\%$ & \textcolor{gray}{$3.54\%$} & $3.08\%$ & $3.24\%$ & $3.58\%$ & $2.56\%$ & $2.52\%$
\\
Cross Sensor~\textcolor{calib3d_red}{$\circ$} & \textcolor{gray}{$4.80\%$} & $4.43\%$ & $4.52\%$ & $4.57\%$ & $4.22\%$ & $4.20\%$ & \textcolor{gray}{$3.27\%$} & $2.83\%$ & $2.96\%$ & $3.36\%$ & $2.81\%$ & $2.78\%$
\\\midrule
Average~\textcolor{calib3d_red}{$\bullet$} & \textcolor{gray}{$6.36\%$} & $5.98\%$ & $6.10\%$ & $6.34\%$ & $5.51\%$ & $5.39\%$ & \textcolor{gray}{$7.57\%$} & $6.94\%$ & $7.14\%$ & $7.50\%$ & $6.19\%$ & $6.03\%$
\\\bottomrule
\end{tabular}}
\vspace{0.2cm}
\end{table*}

\subsection{Comparisons to Recent Calibration Methods}
In the main body of this paper, we provide a comprehensive benchmark study of classical network calibration methods, such as TempS, LogiS, DiriS, and MetaC, across a range of ten different 3D datasets. The benchmark results verify that the proposed DepthS exhibits stronger performance compared to these classical approaches.

To provide a more holistic evaluation of DepthS compared to more recent network calibration methods, we conduct experiments with more recent network calibration methods, including DeepEnsemble \cite{lakshminarayanan2017deepensemble}, BatchEnsemble \cite{wen2020batchensemble}, MIMO \cite{havasi2021mimo}, and PackedEnsemble \cite{laurent2023packedensemble}, on the validation set of the SemanticKITTI \cite{behley2019semanticKITTI} dataset. As shown in \cref{tab:recent_methods}, the results demonstrate that our proposed DeptS is consistently better than both the classical and recent network calibration methods.

\section{Additional Qualitative Result}
\label{sec:supp_qualitative}

In this section, we supplement additional qualitative examples to better support the findings and conclusions drawn in the main body of this paper.

\subsection{Visualized Calibration Results}
\label{subsec:visualized_calib}

\begin{figure*}[t]
    \begin{center}
    \includegraphics[width=0.94\linewidth]{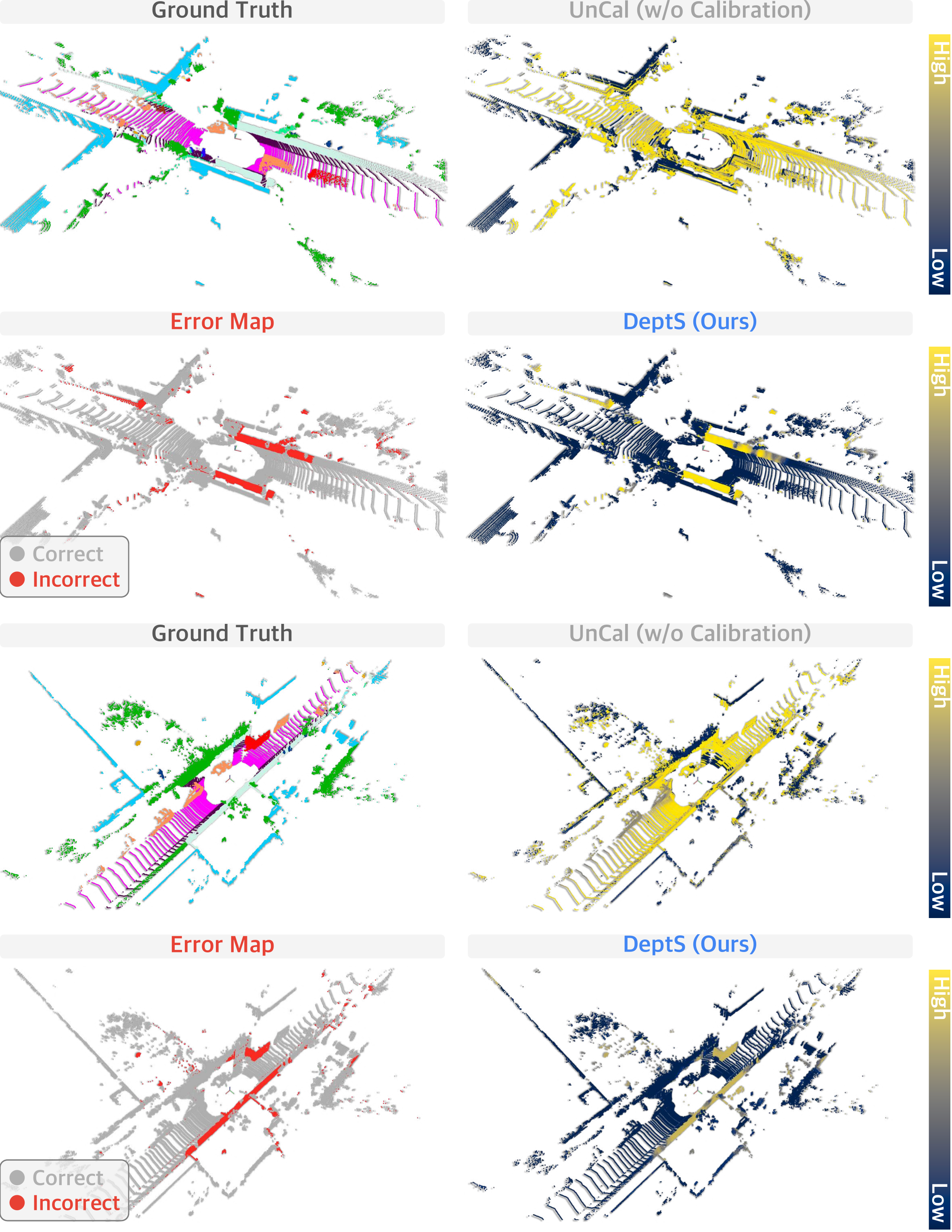}
    \end{center}
    \vspace{-0.5cm}
    \caption{The point-wise expected calibration error (ECE) of existing 3D semantic segmentation models without calibration (\textcolor{calib3d_gray}{UnCal}) and with our depth-aware scaling (\textcolor{calib3d_blue}{DeptS}). Our approach is capable of delivering accurate uncertainty estimates. The colormap goes from \textit{dark} to \textit{light} denotes \textit{low} and \textit{high} error rates, respectively.} 
    \label{fig:supp_semkitti_ece_1}
\end{figure*}

\begin{figure*}[t]
    \begin{center}
    \includegraphics[width=0.94\linewidth]{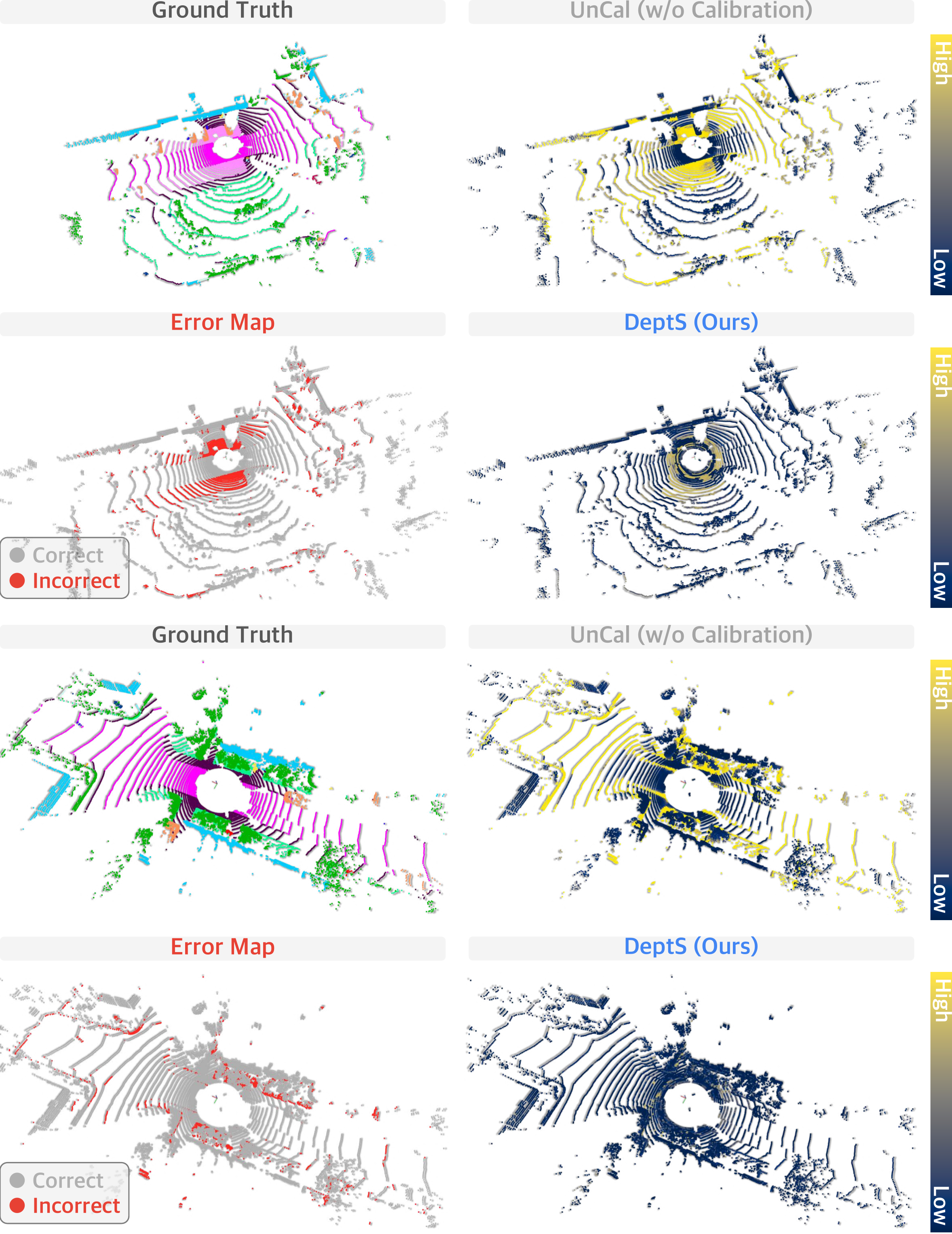}
    \end{center}
    \vspace{-0.5cm}
    \caption{The point-wise expected calibration error (ECE) of existing 3D semantic segmentation models without calibration (\textcolor{calib3d_gray}{UnCal}) and with our depth-aware scaling (\textcolor{calib3d_blue}{DeptS}). Our approach is capable of delivering accurate uncertainty estimates. The colormap goes from \textit{dark} to \textit{light} denotes \textit{low} and \textit{high} error rates, respectively.} 
    \label{fig:supp_semkitti_ece_2}
\end{figure*}

We provide additional visualizations to help verify the effectiveness of the proposed model calibration model in enhancing the model's ability for uncertainty estimation. As can be seen from \cref{fig:supp_semkitti_ece_1} and \cref{fig:supp_semkitti_ece_2}, existing 3D scene understanding models often fail to deliver accurate uncertainty estimates, resulting in potential safety-related issues. Our proposed DeptS is capable of tackling these problems in a holistic manner. After calibration, models can generate more accurate uncertainty estimates, leading to a more reliable 3D scene understanding.

\section{Limitation \& Discussion}
\label{sec:supp_limitation}

In this section, we elaborate on the limitations and potential negative societal impact of this work.

\subsection{Potential Limitations}
\label{subsec:potential_limitation}
In this work, we established the first benchmark of 3D scene understanding from an uncertainty estimation viewpoint. We also proposed DeptS to effectively calibrate 3D models, achieving more reliable 3D scene understanding. We foresee the following limitations that could be promising future directions.

\noindent\textbf{Data Dependence.}
Effective model calibration heavily relies on the quality and diversity of the data used. If the dataset is not representative of real-world scenarios or lacks diversity, the calibrated model may not generalize well across different environments or conditions.

\noindent\textbf{Evaluation Challenges.}
Assessing the effectiveness of calibration can be challenging, as it requires comprehensive metrics that capture the model's performance across a broad range of scenarios. Standard evaluation metrics may not fully reflect the improvements in reliability and confidence achieved through calibration. It is enlightening to design new metrics for a more holistic evaluation.

\subsection{Potential Societal Impact}
\label{subsec:potential_societal_impact}

3D scene understanding often involves capturing and analyzing detailed spatial data about environments which might include private spaces. Additionally, calibrated models might still inherit biases present in the data or algorithmic design, leading to unfair or discriminatory outcomes in certain scenarios. Addressing these issues requires more than technical solutions; it demands careful consideration of the ethical and societal implications of model deployment.

\section{Public Resources Used}
\label{sec:supp_acknowledge}

In this section, we acknowledge the use of the following public resources, during the course of this work.

\subsection{Public Codebase Used}
\label{subsec:acknowledge_codebase}
We acknowledge the use of the following public codebase during this work:
\begin{itemize}
    \item MMCV\footnote{\url{https://github.com/open-mmlab/mmcv}.} \dotfill Apache License 2.0
    \item MMDetection\footnote{\url{https://github.com/open-mmlab/mmdetection}.} \dotfill Apache License 2.0
    \item MMDetection3D\footnote{\url{https://github.com/open-mmlab/mmdetection3d}.} \dotfill Apache License 2.0
    \item MMEngine\footnote{\url{https://github.com/open-mmlab/mmengine}.} \dotfill Apache License 2.0
    \item OpenPCSeg\footnote{\url{https://github.com/PJLab-ADG/OpenPCSeg}.} \dotfill Apache License 2.0
    \item Pointcept\footnote{\url{https://github.com/Pointcept/Pointcept}.} \dotfill MIT License
\end{itemize}

\subsection{Public Datasets Used}
\label{subsec:acknowledge_datasets}

We acknowledge the use of the following public datasets during this work:
\begin{itemize}
    \item nuScenes\footnote{\url{https://www.nuscenes.org/nuscenes}.} \dotfill CC BY-NC-SA 4.0
    \item nuScenes-devkit\footnote{\url{https://github.com/nutonomy/nuscenes-devkit}.} \dotfill Apache License 2.0
    \item SemanticKITTI\footnote{\url{http://semantic-kitti.org}.} \dotfill CC BY-NC-SA 4.0
    \item SemanticKITTI-API\footnote{\url{https://github.com/PRBonn/semantic-kitti-api}.} \dotfill MIT License
    \item WaymoOpenDataset\footnote{\url{https://waymo.com/open}.} \dotfill Waymo Dataset License
    \item SemanticPOSS\footnote{\url{http://www.poss.pku.edu.cn/semanticposs.html}.} \dotfill CC BY-NC-SA 3.0
    \item Synth4D\footnote{\url{https://github.com/saltoricristiano/gipso-sfouda}.} \dotfill GPL-3.0 License
    \item SemanticSTF\footnote{\url{https://github.com/xiaoaoran/SemanticSTF}.} \dotfill CC BY-NC-SA 4.0
    \item ScribbleKITTI\footnote{\url{https://github.com/ouenal/scribblekitti}.} \dotfill Unknown
    \item S3DIS\footnote{\url{http://buildingparser.stanford.edu/dataset.html}.} \dotfill Unknown
    \item Robo3D\footnote{\url{https://github.com/ldkong1205/Robo3D}.} \dotfill CC BY-NC-SA 4.0
\end{itemize}

\subsection{Public Implementations Used}
\label{subsec:acknowledge_implements}

We acknowledge the use of the following implementations during this work:
\begin{itemize}
    \item lidar-bonnetal\footnote{\url{https://github.com/PRBonn/lidar-bonnetal}.} \dotfill MIT License
    \item SalsaNext\footnote{\url{https://github.com/TiagoCortinhal/SalsaNext}.} \dotfill MIT License
    \item FIDNet\footnote{\url{https://github.com/placeforyiming/IROS21-FIDNet-SemanticKITTI}.} \dotfill Unknown
    \item CENet\footnote{\url{https://github.com/huixiancheng/CENet}.} \dotfill MIT License
    \item rangevit\footnote{\url{https://github.com/valeoai/rangevit}.} \dotfill Apache License 2.0
    \item FRNet\footnote{\url{https://github.com/Xiangxu-0103/FRNet}.} \dotfill Apache License 2.0
    \item PolarSeg\footnote{\url{https://github.com/edwardzhou130/PolarSeg}.} \dotfill BSD 3-Clause License
    \item MinkowskiEngine\footnote{\url{https://github.com/NVIDIA/MinkowskiEngine}.} \dotfill MIT License
    \item TorchSparse\footnote{\url{https://github.com/mit-han-lab/torchsparse}.} \dotfill MIT License
    \item SPVNAS\footnote{\url{https://github.com/mit-han-lab/spvnas}.} \dotfill MIT License
    \item Cylinder3D\footnote{\url{https://github.com/xinge008/Cylinder3D}.} \dotfill Apache License 2.0
    \item spconv\footnote{\url{https://github.com/traveller59/spconv}.} \dotfill Apache License 2.0
    \item 2DPASS\footnote{\url{https://github.com/yanx27/2DPASS}.} \dotfill MIT License
    \item CPGNet\footnote{\url{https://github.com/GangZhang842/CPGNet}.} \dotfill Unknown
    \item GFNet\footnote{\url{https://github.com/haibo-qiu/GFNet}.} \dotfill Unknown
    \item KPConv\footnote{\url{https://github.com/HuguesTHOMAS/KPConv}.} \dotfill MIT License
    \item PIDS\footnote{\url{https://github.com/lordzth666/WACV23_PIDS-Joint-Point-Interaction-Dimension-Search-for-3D-Point-Cloud}.} \dotfill MIT License
    \item PointTransformerV2\footnote{\url{https://github.com/Pointcept/PointTransformerV2}.} \dotfill Unknown
    \item WaffleIron\footnote{\url{https://github.com/valeoai/WaffleIron}.} \dotfill Apache License 2.0
    \item selectivecal\footnote{\url{https://github.com/dwang181/selectivecal}.} \dotfill Unknown
    \item LaserMix\footnote{\url{https://github.com/ldkong1205/LaserMix}.} \dotfill CC BY-NC-SA 4.0
    \item PolarMix\footnote{\url{https://github.com/xiaoaoran/polarmix}.} \dotfill MIT License
\end{itemize}

%%%%%%%%% REFERENCES
\clearpage
{\small
\bibliographystyle{ieee_fullname}
\bibliography{egbib}
}

\end{document}